\documentclass{article}
\usepackage[utf8]{inputenc}

\usepackage{times}
\usepackage{epsfig}
\usepackage{graphicx}
\usepackage{subcaption}
\usepackage{float}
\usepackage{amsmath}
\usepackage{amssymb}
\usepackage{enumitem}
\usepackage{hyperref}

\begin{document}

%%%%%%%%% TITLE
\title{Image-Guided Depth Sampling and Reconstruction}

\author{Adam Wolff, Shachar Praisler, Ilya Tcenov and Guy Gilboa\\
Technion - Israel Institute of Technology\\
Haifa, Israel\\
{\tt\small \{sadamwol,spraizler,ilya.tcenov\}@campus.technion.ac.il,}\\ {\tt\small guy.gilboa@ee.technion.ac.il}
% \and
% Shachar Praisler\\
% Technion - Israel Institute of Technology\\
% Haifa, Israel\\
% {\tt\small spraizler@campus.technion.ac.il}
% \and
% Ilya Tcenov\\
% Technion - Israel Institute of Technology\\
% Haifa, Israel\\
% {\tt\small ilya.tcenov@campus.technion.ac.il}
% \and
% Guy Gilboa\\
% Technion - Israel Institute of Technology\\
% Haifa, Israel\\
% {\tt\small guy.gilboa@ee.technion.ac.il}
}

\maketitle

%%%%%%%%% ABSTRACT
\begin{abstract}
Depth acquisition, based on active illumination, is essential for autonomous and robotic navigation. LiDARs (Light Detection And Ranging) with mechanical, fixed, sampling templates are commonly used in today's autonomous vehicles. An emerging technology, based on solid-state depth sensors, with no mechanical parts, allows fast, adaptive, programmable scans. 

In this paper, we investigate the topic of adaptive, image-driven, sampling and reconstruction strategies. First, we formulate a piece-wise linear depth model with several tolerance parameters and estimate its validity for indoor and outdoor scenes. Our model and experiments predict that, in the optimal case, about 20-60 piece-wise linear structures can approximate well a depth map. This translates to a depth-to-image sampling ratio of about 1/1200. We propose a simple, generic, sampling and reconstruction algorithm, based on super-pixels. We reach a sampling rate which is still far from the optimal case. However, our sampling improves grid and random sampling, consistently, for a wide variety of reconstruction methods. Moreover, our proposed reconstruction achieves state-of-the-art results, compared to image-guided depth completion algorithms, reducing the required sampling rate by a factor of 3-4.
A single-pixel depth camera built in our lab illustrates the concept.\footnote{A video demonstration of the device is available at: \url{https://youtu.be/7_DDCXL25uE}}
\end{abstract}

\graphicspath{{./figures/teaser/}}
\begin{figure}[t!]
    \centering
    \begin{subfigure}{0.45\textwidth}
        \centering
        \includegraphics[height=1.2in]{rgb.png} 
       \caption{RGB}
    \end{subfigure}
    \hspace{1em}
    \begin{subfigure}{0.45\textwidth}
        \centering
        \includegraphics[height=1.2in]{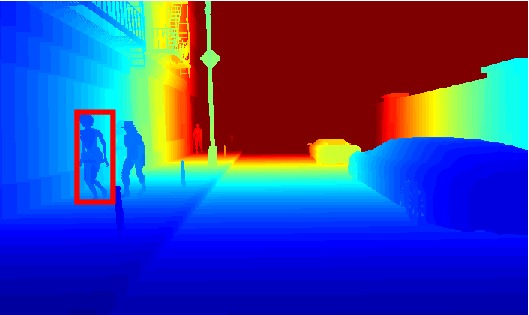}
        \caption{Ground-truth depth}
        \label{Fig::Ground truth}
    \end{subfigure}
    \\
    \vspace{10pt}
    \begin{subfigure}{0.3\textwidth}
        \centering
        \includegraphics[height=2.0in]{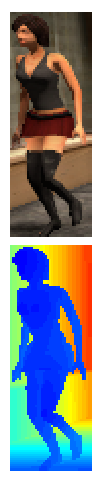}
        \caption{Patch RGB+GT}
    \end{subfigure}
    % \hspace{1em}
    \begin{subfigure}{0.2\textwidth}
        \centering
        \includegraphics[height=2.0in]{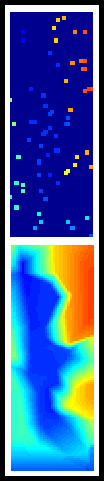}
        \caption{}
    \end{subfigure}
    \begin{subfigure}{0.2\textwidth}
        \centering
        \includegraphics[height=2.0in]{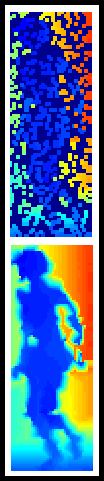}
        \caption{}
    \end{subfigure}
    \begin{subfigure}{0.2\textwidth}
        \centering
        \includegraphics[height=2.0in]{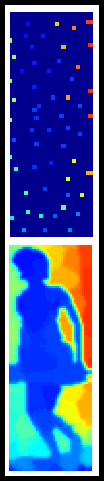}
        \caption{}
    \end{subfigure}
    \caption{We developed an algorithm for RGB-based depth sampling and reconstruction. Our method focuses mostly on small and thin objects. While using 66 random samples followed by bilinear interpolation provides poor result (d) with RMSE=0.52m within patch area, our method reconstructs the pedestrian almost perfectly with only 57 samples (f) with RMSE=0.31m. To achieve the same RMSE, bilinear interpolation requires 669 samples (e).}
    \label{fig:teaser}
\end{figure}

%%%%%%%%% BODY TEXT
\section{Introduction}
In recent years, depth sensing has become essential for a variety of new significant applications. For example, depth sensors assist autonomous cars in navigation and in collision prevention \cite{schwarz2010lidar}. The physical constraints on active depth sensing mobile devices, such as light detection and ranging (LiDAR), yield sparse depth measurements per scan. This results in a coarse point cloud and requires an additional estimation of missing data. %The accuracy of the estimation process relies not only on the chosen depth prediction method but also on the spatial distribution of samples. 

Traditional LiDARs have a restricted scanning mechanism. Those devices measure distance in specified angle intervals, using a fixed number of horizontal scan-lines (usually 16 to 64), depending on the number of transceivers. A new revolutionary technology is now emerging of solid-state depth sensors. They are based on optical phased-arrays with no mechanical parts, and can thus scan the scene fast in an adaptive manner (programmable scanning) \cite{cheben2018subwavelength,poulton2017coherent}. In addition, those innovative devices are much cheaper than those currently in use. This calls for the development of new, efficient, sampling strategies, which reduce the reconstruction error per sample. Since almost always autonomous platforms are equipped with RGB cameras, we investigate the possibility to improve the depth sampling process by taking the RGB information into account.

In this paper, we address the topic of image-guided depth sampling and reconstruction. First, we introduce the concept of adaptive depth sampling and develop an appropriate model of the data. Then, we introduce a fast and practical image-guided algorithm for depth sampling and reconstruction, based on super-pixels. An example of output of a our algorithm is shown in Fig. \ref{fig:teaser}.
We demonstrate in experiments that our framework outperforms state-of-the-art depth completion methods for both indoor and outdoor scenes. Finally, since current solid-state technology is not yet technically-open for reconfiguration of the sampling, we illustrate the concept in real life, by a single-pixel depth camera, which was 3D-printed in our lab. 

%-------------------------------------------------------------------------
\section{Related Work}\label{Sec::related work}
{\bf Depth completion:}
The task of depth reconstruction from scattered sparse samples is being increasingly investigated. The main methods can be divided to those which require only the sparse depth input (unguided) and to those assisted by additional information, e.g. color image (guided). 

Among the unguided methods, some use classical approach \cite{ku2018defense,ma2017sparse}, while others rely on more advanced tools such as deep learning \cite{chodosh2018deep,uhrig2017sparsity}.

On the contrary, guided methods exploit the connection between depth maps and their corresponding color image. Earlier methods used traditional image processing tools \cite{barron2016fast, drozdov2016robust}. Recently, several deep learning-based methods \cite{chen2018estimating,eldesokey2018propagating,huang2018hms,jaritz2018sparse,li2018depth,liao2017parse,ma2018self,ma2018sparse} achieved state-of-the-art results.

% \cite{barron2016fast} considered the problem of depth superresolution through image filtering. Drozdov \etal \cite{drozdov2016robust} used super-pixel technique, followed by the total generalized variation (TGV) regularization.

% Ku \etal \cite{ku2018defense} developed a simple and fast algorithm, relying on basic image processing operations. \cite{liu2015depth} introduced a method based on the alternating direction method of multipliers (ADMM), using a combined wavelet-contourlet dictionary for depth map representation. \cite{zeglazi2018efficient} proposed a stereo matching algorithm, starting with sampling boundary pixels in super-pixel map, and later reconstructing the disparity map by a scanline propagation method. 

{\bf Early guided depth sampling:}
Despite the intensive development in depth completion, the issue of adaptive sampling is yet little addressed. Only \cite{hawe2011dense,liu2015depth} have offered a non-trivial (i.e. uniformly random or grid) sampling pattern as a previous step to depth reconstruction. Both studies selected sampling at locations which are most probable to have strong depth gradient. Nonetheless, they failed dealing with very low sampling budget of less than 5\% of ground-truth pixels.

{\bf Nonuniform sampling:}
Over the years, the field of nonuniform sampling has been well established \cite{aldroubi2001nonuniform,babu2010spectral,marvasti2012nonuniform,yen1956nonuniform}. However, these studies focus on the reconstruction of the signal for a given nonuniform sampling pattern and not on how to design data-driven patterns, given side information. 
%-------------------------------------------------------------------------
\section{The Space of Depth Images}
Any sampling strategy is based on a model of the signal to be sampled. For example, in classical Fourier analysis, the assumption is of band-limited signals. Thus, sampling at the Nyquist frequency guarantees perfect reconstruction. 
Compressed sensing \cite{candes_tao2006} assumes a sparse underlying model of the signal (such as in terms of edges). Sub-Nyquist sampling \cite{eldar}, is based on the ability to manipulate correctly aliased signals, based on prior knowledge of the frequency structure of the data. However, the models above assume a single source of data to be sampled. We would like to examine an appropriate model for depth scenes, as well as the relation to the RGB data of the same scene. 
We propose a simple depth model and try to validate it experimentally on benchmark data. We then relate it to RGB.

\graphicspath{{./figures/motivation/}}
% depth model - example
\begin{figure}[htb]
    \centering
    \begin{subfigure}{0.15\textwidth}
        \hspace{0.07pt}
    \end{subfigure}
    \begin{subfigure}{0.40\textwidth}
        \centering
        Synthia
    \end{subfigure}
    \begin{subfigure}{0.35\textwidth}
        \centering
        NYU-Depth-v2
    \end{subfigure}
    \\
    \begin{subfigure}{0.15\textwidth}
        \centering
        RGB
    \end{subfigure}
    \begin{subfigure}{0.40\textwidth}
        \centering
        \includegraphics[height=1.1in]{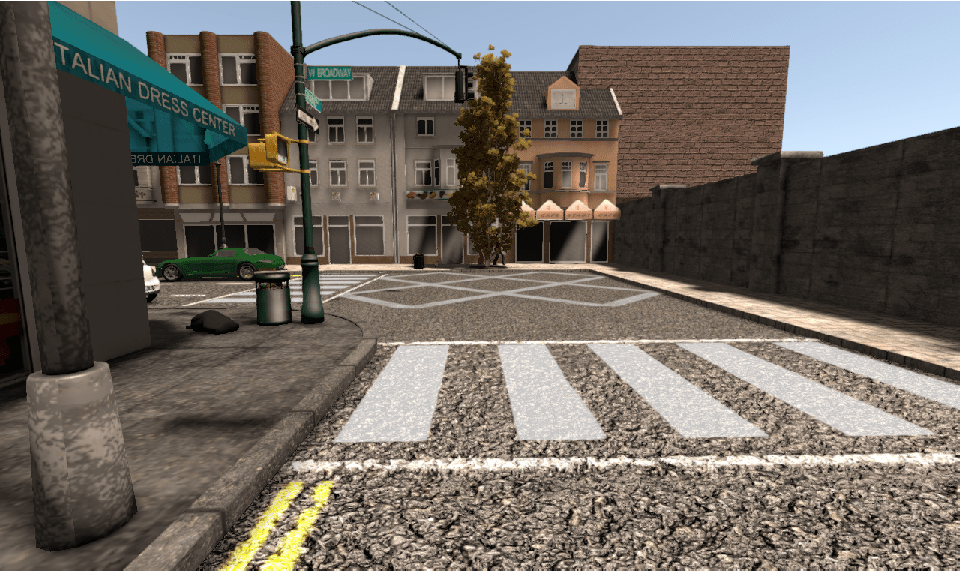}
    \end{subfigure}
    \begin{subfigure}{0.35\textwidth}
        \centering
        \includegraphics[height=1.1in]{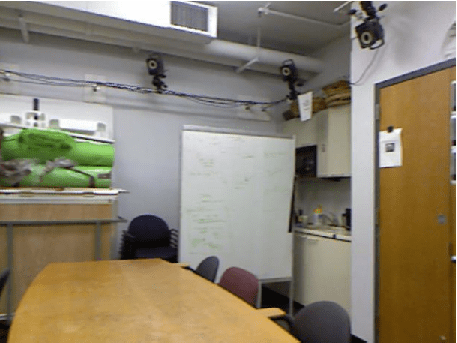} 
    \end{subfigure}
    \\
    \begin{subfigure}{0.15\textwidth}
        \centering
        GT
    \end{subfigure}
    \begin{subfigure}{0.40\textwidth}
        \centering
        \includegraphics[height=1.1in]{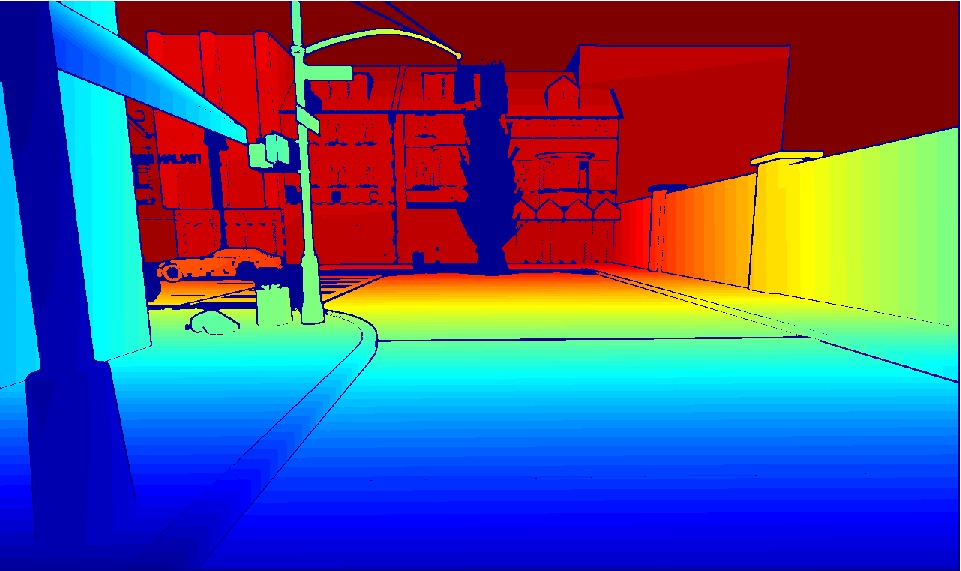} 
    \end{subfigure}
    \begin{subfigure}{0.35\textwidth}
        \centering
        \includegraphics[height=1.1in]{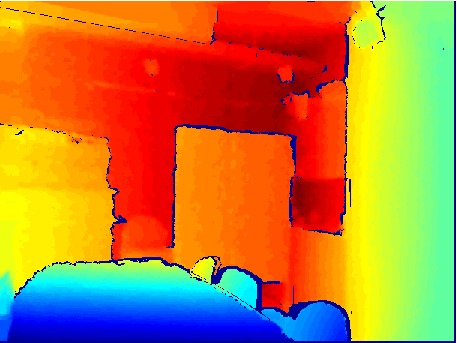} 
    \end{subfigure}
    \\
    \begin{subfigure}{0.15\textwidth}
        \centering
        Piece-wise planar approx.
    \end{subfigure}
    \begin{subfigure}{0.40\textwidth}
        \centering
        \includegraphics[height=1.1in]{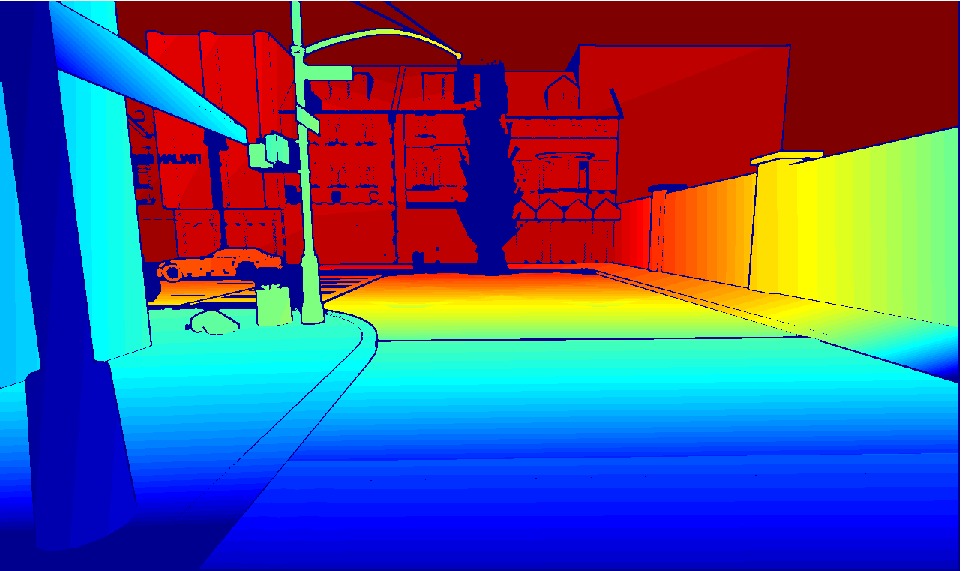} 
    \end{subfigure}
    \begin{subfigure}{0.35\textwidth}
        \centering
        \includegraphics[height=1.1in]{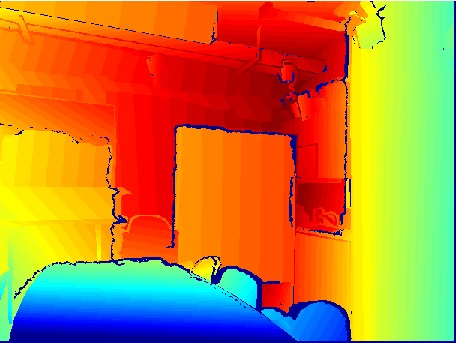} 
    \end{subfigure}
    \caption{Examples for piece-wise planar approximation.}
    \label{fig:model_example}    
\end{figure}

\graphicspath{{./figures/motivation/}}
% depth model - statistics
\begin{figure}[htb]
    \centering
    \begin{subfigure}{0.45\textwidth}
        \centering
        Synthia
    \end{subfigure}
    \begin{subfigure}{0.45\textwidth}
        \centering
        NYU-Depth-v2
    \end{subfigure}
    \\
    \begin{subfigure}{0.45\textwidth}
        \centering
        \includegraphics[height=1.3in]{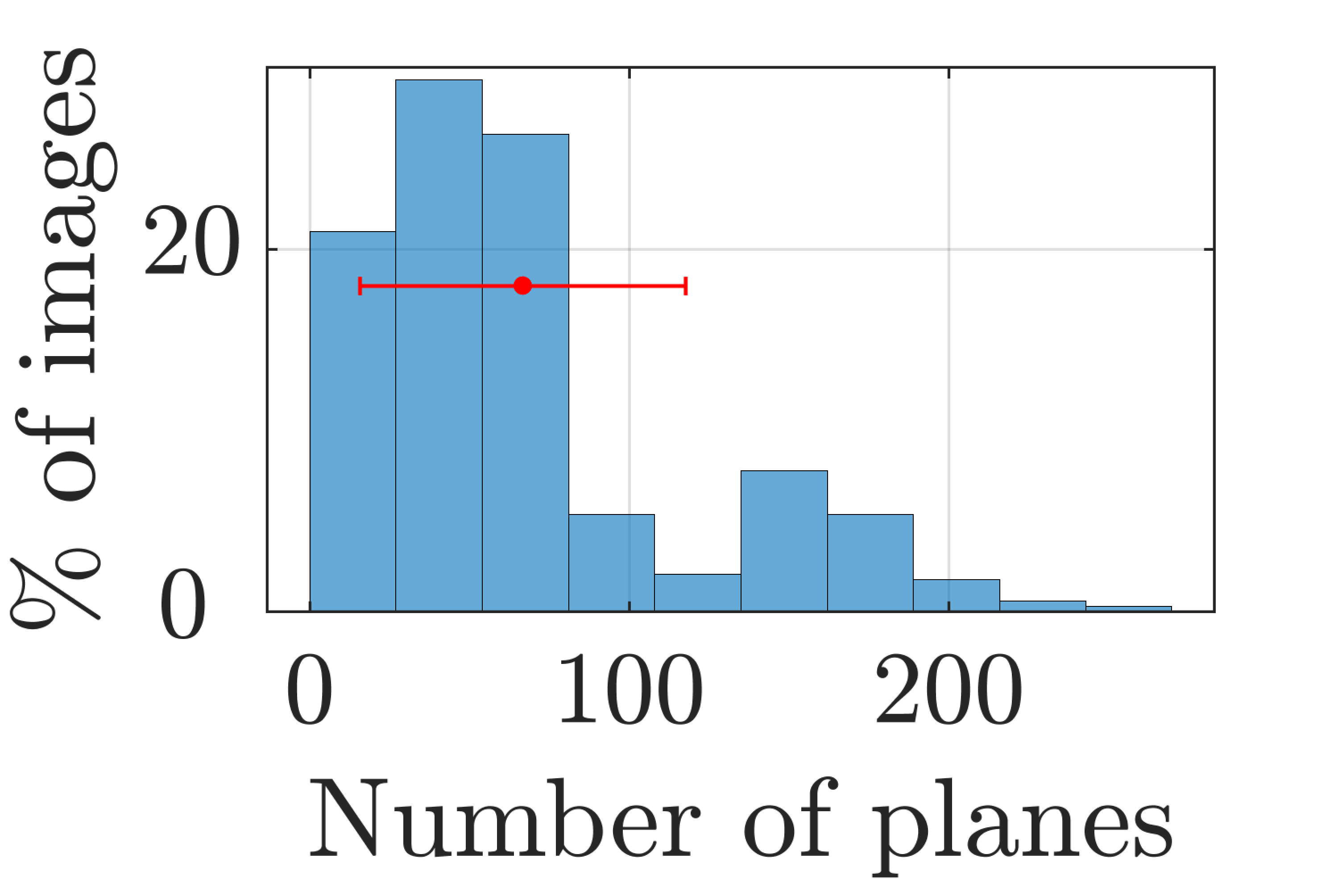}
    \end{subfigure}
    \begin{subfigure}{0.45\textwidth}
        \centering
        \includegraphics[height=1.3in]{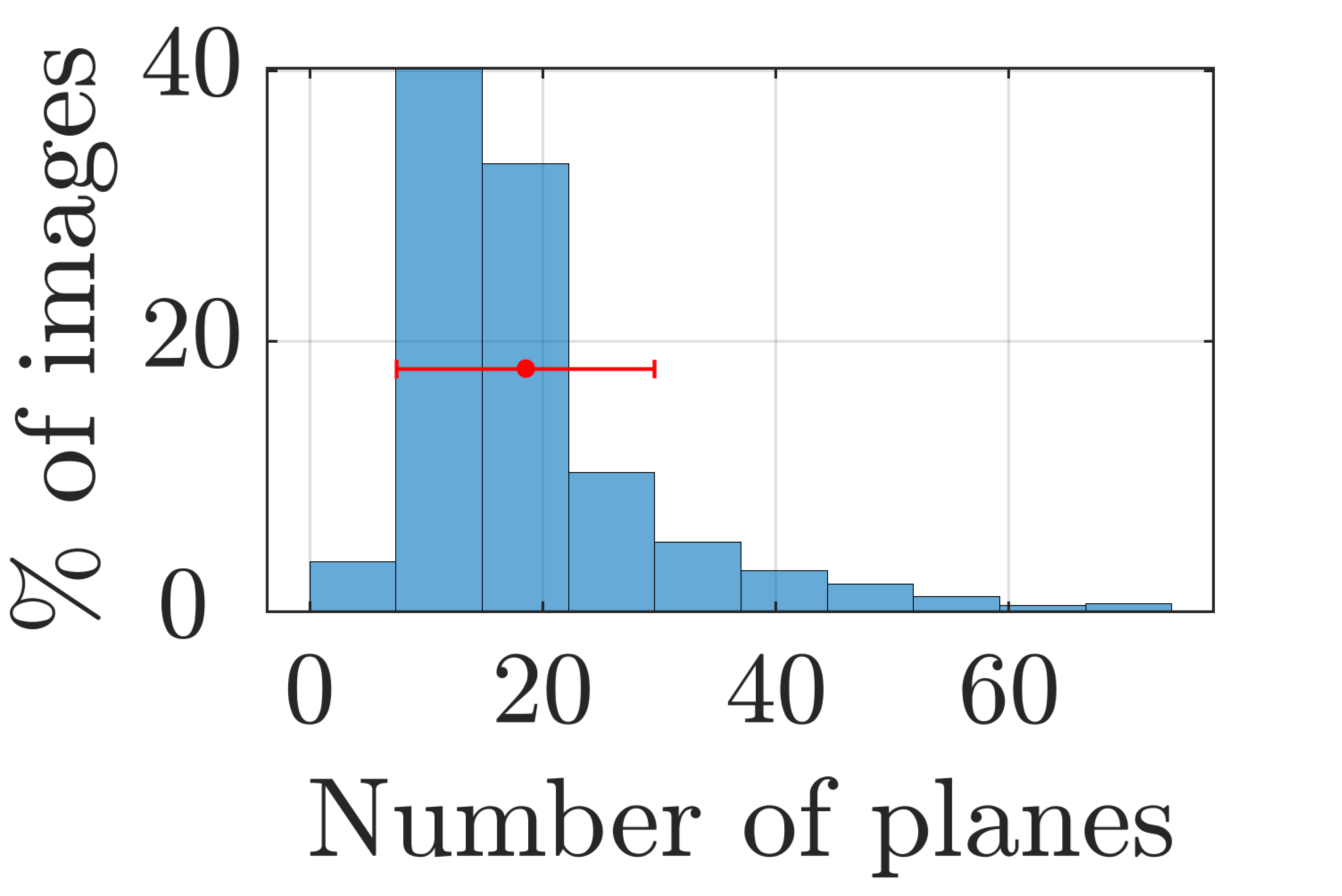} 
    \end{subfigure}
    \\
    \begin{subfigure}{0.45\textwidth}
        \centering
        \includegraphics[height=1.3in]{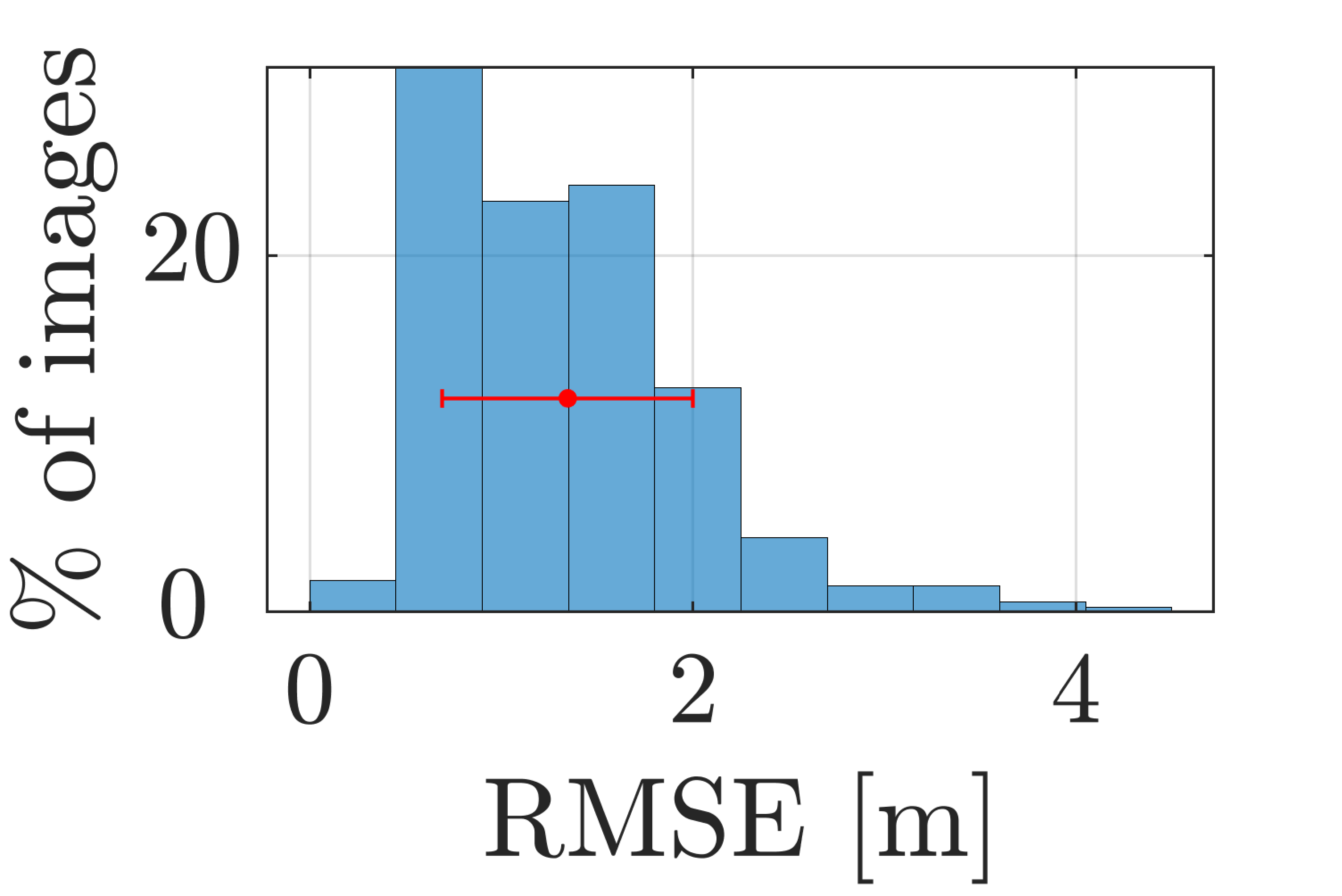} 
    \end{subfigure}
    \begin{subfigure}{0.45\textwidth}
        \centering
        \includegraphics[height=1.3in]{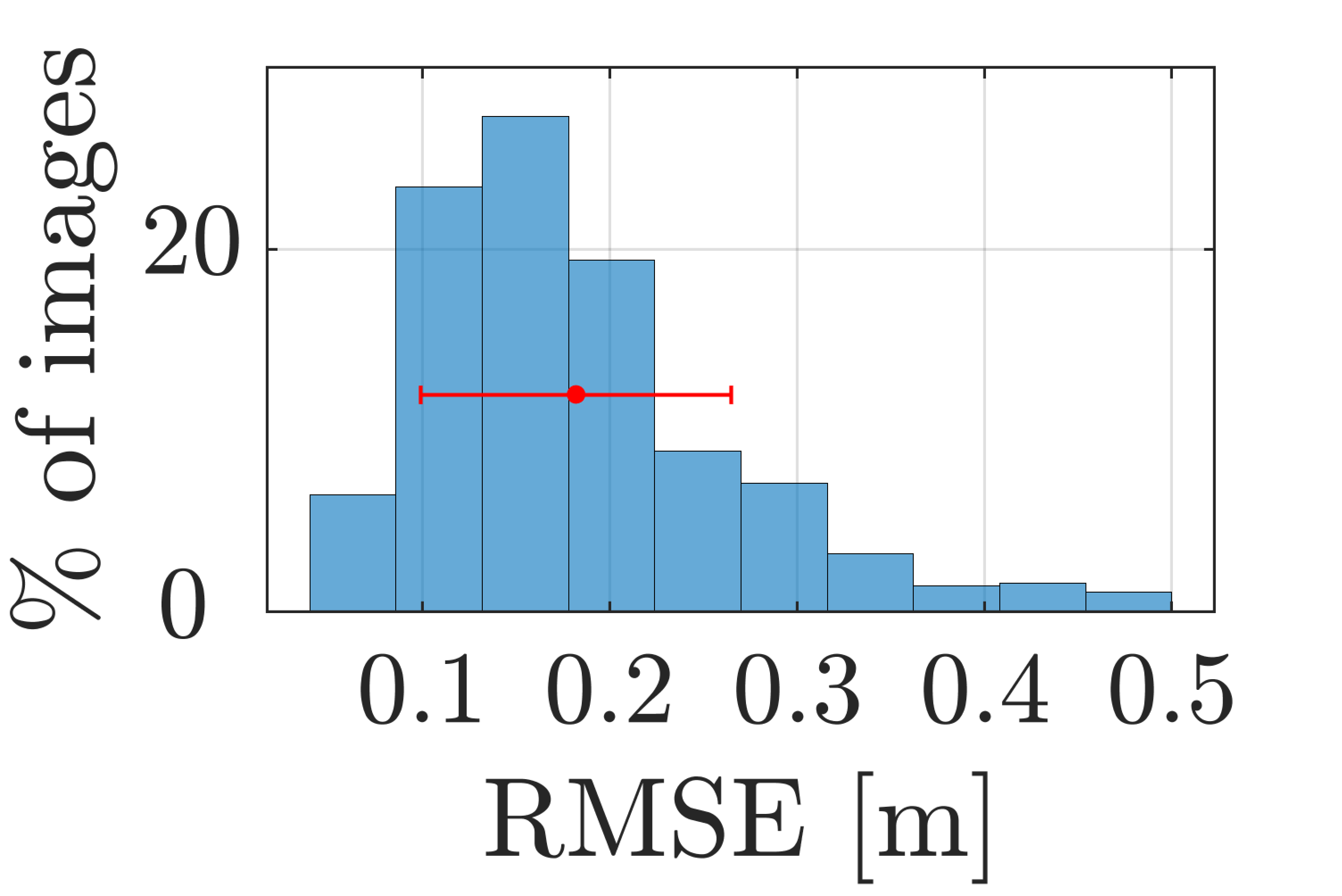} 
    \end{subfigure}
    \caption{Statistics of piece-wise planar approximation. Top: percentage of approximated planes $N$ per image. Bottom: $RMSE_v(d,\hat{d})$, as defined in Eq. \eqref{eq:depth_model}.}
    \label{fig:model_stats}    
\end{figure}

% depth model failures & success - Synthia & NYUv2
\begin{figure}[htb]
    \centering
    \begin{subfigure}{0.95\textwidth}
        \centering
        \includegraphics[width=0.95\textwidth]{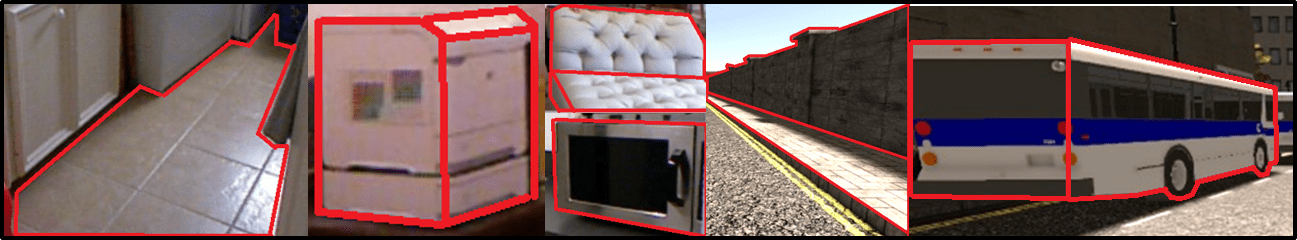} 
        \caption{Fit: piece-wise approximately flat objects}
    \end{subfigure}
    \\
    \vspace{3pt}
    \begin{subfigure}{0.95\textwidth}
        \centering
        \includegraphics[width=0.95\textwidth]{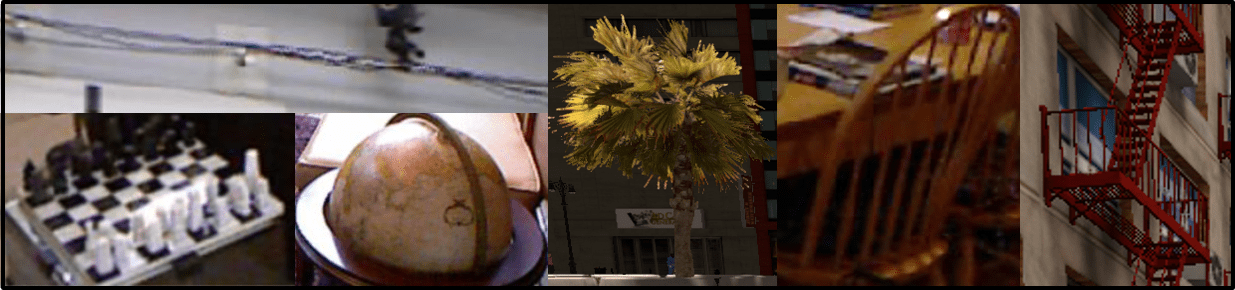} 
        \caption{Do not fit: non flat and non convex objects}
    \end{subfigure}
    \caption{Examples of image parts which fit (top) and parts which do not fit (bottom) the piece-wise planar depth model.}
    \label{fig:failures_success}    
\end{figure}

\subsection{Piece-wise planar depth model}
\label{model_section}
Our primary objective is to obtain depth information for autonomous navigation. Thus, an appropriate model should represent well the general geometrical setting (roads, walls, sidewalks) as well as the location of significant landmarks and obstacles (poles, signs, rocks and objects in a room). For objects, we would like to obtain their location but not necessarily their precise geometry. This leads us to a piece-wise planar model, which was mentioned in \cite{baker1998layered,tao2001global} but not yet formulated and tested.
Given a depth image $d$ our hypothesis is that most of the scene can be well represented by a piece-wise planar approximation.  More formally, let  $\Omega \subset \mathbb{R}^2$ be the image domain, where $|\Omega|$ is its area.
%$\partial \Omega$ is its boundary and $|\partial \Omega|$ is the length of the boundary. 
Let $E=\{\Omega_i\}$, $i=1,..,N$, be a set of $N$ sub-domains which define a partition of $\Omega$. Thus, $\Omega_i \subset \Omega$, $\Omega_i \cap \Omega_j = \emptyset$, $\forall i\ne j$, $\bigcup_{i=1..N} \Omega_i=\Omega$. Let  $\hat{d}$ be a 2D piece-wise linear function, defined on the domain $\Omega$ by
\begin{equation}
    \label{eq:d_hat}
    \hat{d}(x,y) = a_i x + b_i y + c_i, \quad (x,y)\in \Omega_i,\,\, i=1..N,
\end{equation}
where $a_i,b_i,c_i$ are some constants. Let $v$ be a binary function in $\Omega$ which indicates validity of the model, where $1$ indicates validity and $0$ invalidity. We denote by $V$ the set of valid points, $V=\{(x,y)\,|\,v(x,y)=1 \}$. We assume $|V| \ge (1 - \delta)|\Omega|$, $0 \le \delta \ll 1$. Our hypothesis is that given some small tolerance parameters $\delta,\, \epsilon$, a validity map $v$ and a small number of regions $N$, for any depth map $d$ there exists a piece-wise planar approximation, defined by Eq. \eqref{eq:d_hat}, such that
\begin{equation}
    \label{eq:depth_model}
    RMSE_v(d,\hat{d}) \equiv \sqrt{\tfrac{1}{|V|}\| v \cdot (d - \hat{d}) \|_2^2} \le \epsilon.
\end{equation}
Thus, $d$ can be well approximated by a 2D piece-wise linear function, almost everywhere, provided we know the partition set $E$ and the plane parameters $a_i,b_i,c_i$ for each $\Omega_i$. Our aim is to approximate $E$ from the RGB image. In order to recover $\hat{d}$ we need to sample each region $\Omega_i$ 3 times, to estimate its coefficients (in the noiseless case). This gives us a lower bound on the number of samples required to obtain a high quality depth image:
\begin{equation}
    \label{eq:num_samples}
    n_{min} \equiv Number-of-Samples \ge 3N. 
\end{equation}
We now turn to experimentally check this hypothesis and examine the values of $\delta$, $\epsilon$ and $N$ in indoor and outdoor scenes.

To validate the proposed model, we made a piece-wise planar depth approximation for two datasets which have dense ground-truth depth. 
For outdoor scenes, there are little real-life benchmarks with dense depth, we therefore resorted to a high quality emulation, using 787 images downsampled to $640\times380$ pixels in summer sequence 5 (left stereo, front view) of Synthia \cite{ros2016synthia} dataset. 
For indoor scenes we used 654 images downsampled and center-cropped to $304\times228$ pixels (persisting \cite{ma2018sparse}) of NYU-Depth-v2 \cite{silberman2012indoor} test set.
% The specifics of the approximation algorithm are provided in the supplementary material. 

Examples of piece-wise planar approximations are shown in Fig. \ref{fig:model_example} (bottom), compared to the ground truth (middle row). Dark-blue indicates regions not in the set $V$. It can be observed that the approximation is quite accurate.
Statistical results are presented in Fig. \ref{fig:model_stats}. 
The average model parameters recovered for Synthia are $N=66.6$, $\delta=0.1$, $\epsilon = 1.35m$. 
The average model parameters recovered for NYU-Depth-v2 are  $N=18.5$, $\delta=0.07$, $\epsilon = 0.18m$.

In these cases, according to Eq. \eqref{eq:num_samples}, Synthia can be well approximated in an optimal scenario by an average of only 200 samples, whereas NYU-v2 by an average of 56 samples (in both cases, this translates to about $0.08\%$ sampling ratio, compared to the ground-truth depth resolution). In Fig. \ref{fig:failures_success} we show examples of objects which fit well a piece-wise planar approximation (top) and counter examples of highly non-convex structures or ones with high curvature.

%\subsubsection{Small objects approximation}
%Settle with explanation in text, refer to later plots of 0 vs. 1st order. \textcolor{red}{Maybe we should discuss it at the method section, where we compare 0-order to 1st-order (although we evaluate on the complete image and not on obstacles).}

\subsection{Relation of RGB and depth}
\begin{figure}[htb]
    \centering
    \begin{subfigure}{0.5\textwidth}
        \centering
        \frame{\includegraphics[height=1.3in]{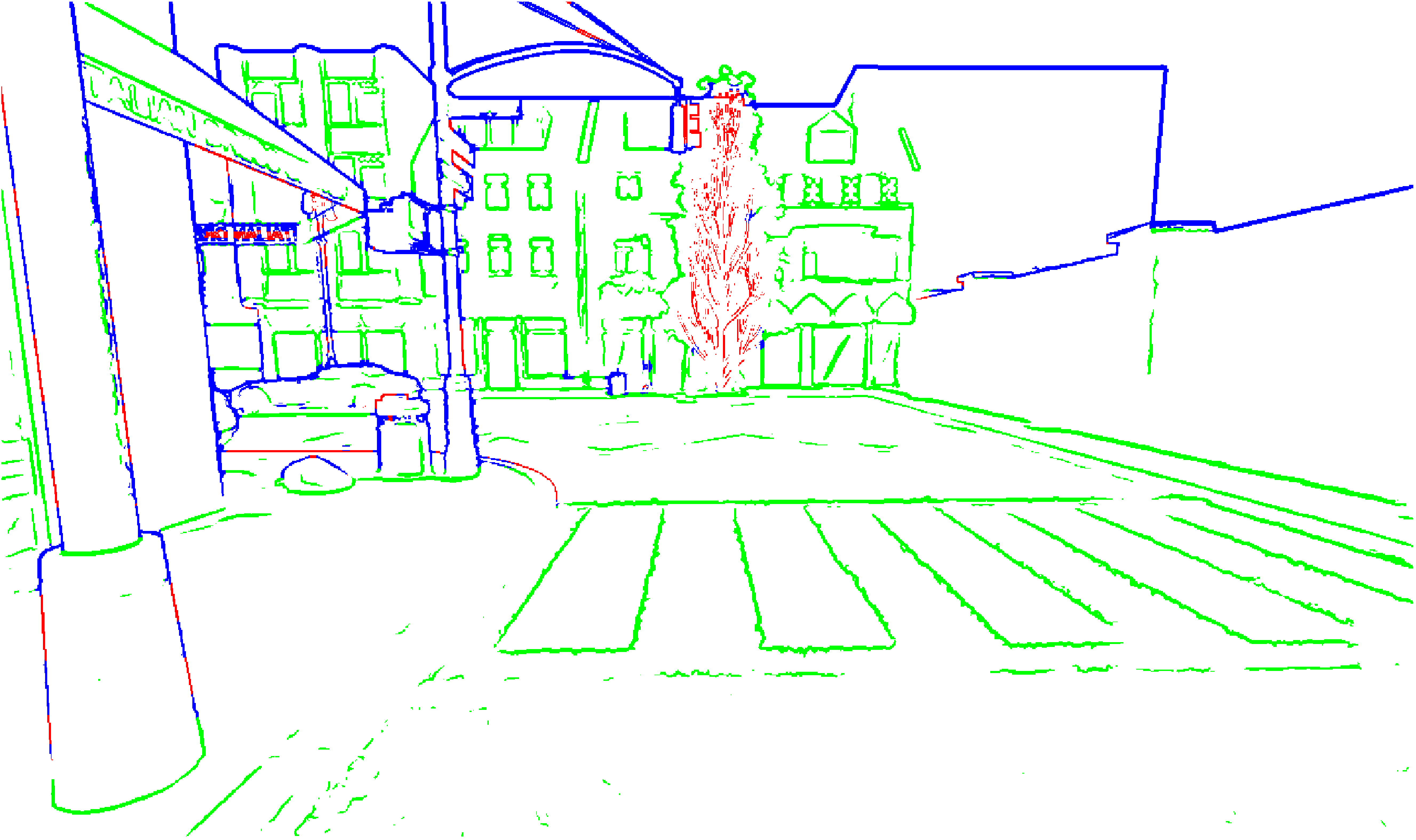}}
    \end{subfigure}
    \begin{subfigure}{0.45\textwidth}
        \centering
        \frame{\includegraphics[height=1.3in]{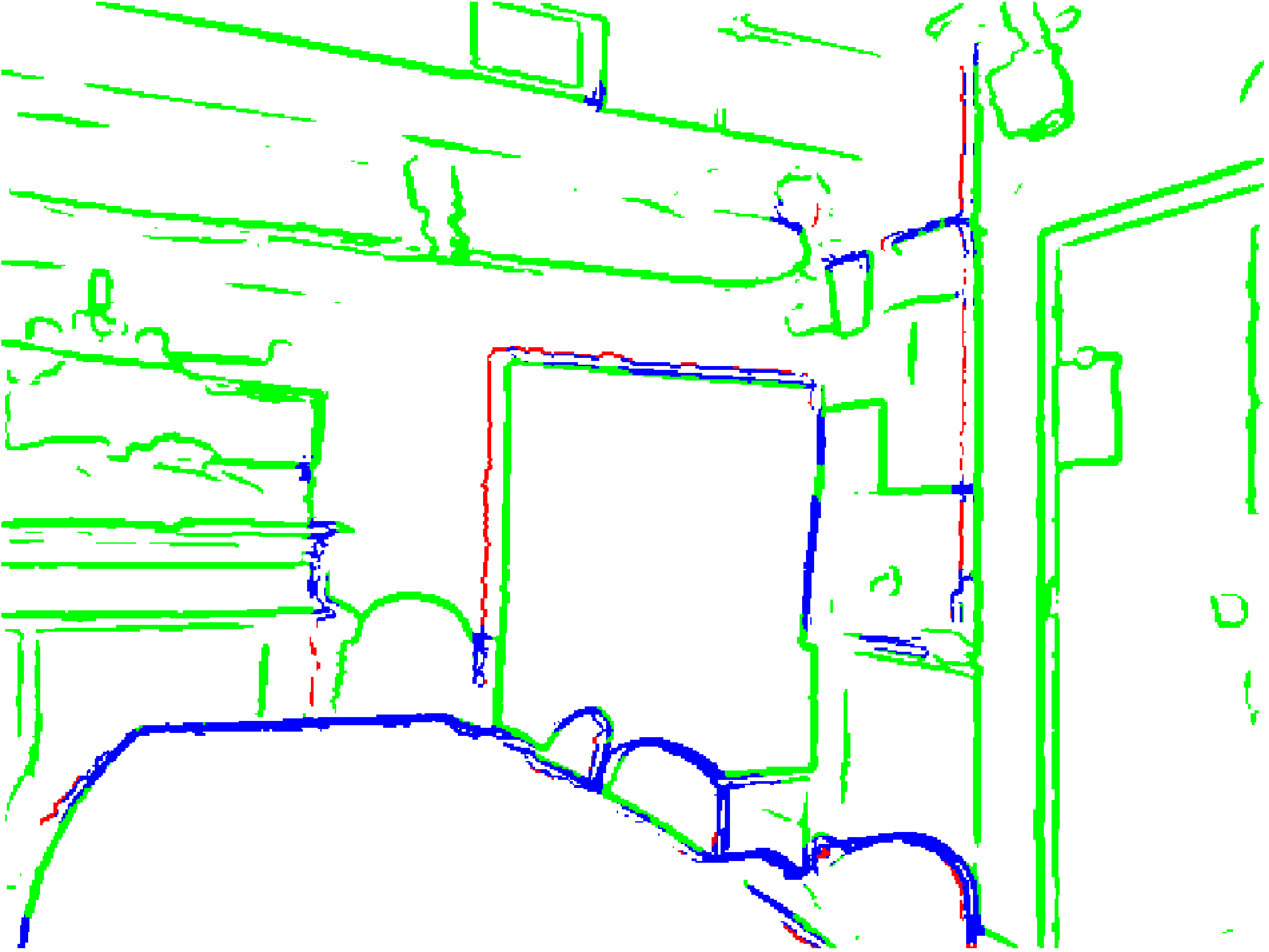}}
    \end{subfigure}
    \caption{Examples of depth discontinuities (red) and RGB edges (green) correlation. Blue pixels include both types.}
    \label{fig:dc_edges_qual}    
\end{figure}

Next, we want to examine the possibility to estimate the partition set $E$ from the RGB data. This is a very challenging task, which is an open problem at this point.
We thus turn to a simpler problem of checking the relation between RGB edges and depth discontinuities. 
Given the set of RGB boundaries (edges) $B_{rgb} \subset \Omega$ and depth boundaries $B_{d} \subset \Omega$ we would like to calculate empirically, for each coordinate ${\bf x}=(x,y)\in \Omega$, the following conditional probabilities:
\begin{align}
P_{rgb-d} &= Prob({\bf x}\in B_{rgb}\,|\,{\bf x}\in B_{d}),\\
P_{d-rgb} &= Prob({\bf x}\in B_{d}\,|\,{\bf x}\in B_{rgb}).
\end{align}
We compute the set $B_{rgb}$ for each image by using a generic edge-detector, well suited for natural images \cite{dollar2013structured}. For the set $B_{d}$ we employed a threshold on the depth gradient, normalized by the depth value.
% More details can be found in the supplementary material. 
We allowed some tolerance in the registration of the images due to misalignment, so for any ${\bf x}$, the search is in a $5 \times 5$ pixel neighborhood.
For Synthia we got $P_{rgb-d}=69.3\%$, $P_{d-rgb}=33.7\%$.
For NYU-v2 we got $P_{rgb-d}=80.8\%$, $P_{d-rgb}=22.9\%$.
The high values of $P_{rgb-d}$ indicate the ability to predict well depth discontinuities, based on RGB edges. The relatively low values of $P_{d-rgb}$ indicate that we should expect many false partitions (which appear only in the RGB, but not in the depth data).  Thus we can expect to be able to approximate, to some extent, the partition set $E$ based solely on the RGB image, by over-segmentation. As the partition is quite a rough approximation, additional samples are required, above the lower bound expressed in Eq. \eqref{eq:num_samples}. In Fig. \ref{fig:dc_edges_qual} we show examples of boundaries in the RGB and depth for both sets.

% depth discontinuities vs. color edges

%-------------------------------------------------------------------------

\section{Method}
We propose a generic and simple method for depth sparse sampling and dense reconstruction.
The following assumptions are made:
\begin{enumerate}
    \item Measurements are of high quality, such that noise of the range measurement is negligible, compared to the global error of the dense reconstruction. 
    \item Sampling budget is limited to $n$ samples.
    \item An RGB image of the scene is available to guide the process. The reconstructed depth is registered to this image. Sensitive cameras may be used for night scenes.
    \item Sampling is point-wise. The system can sample at any desired location of the RGB image. The sampling pattern can change for each image.
\end{enumerate}

% Algorithm figure
\graphicspath{{./figures/algorithm/}}
\begin{figure*}[htb]
    \centering
    \includegraphics[width=0.95\textwidth]{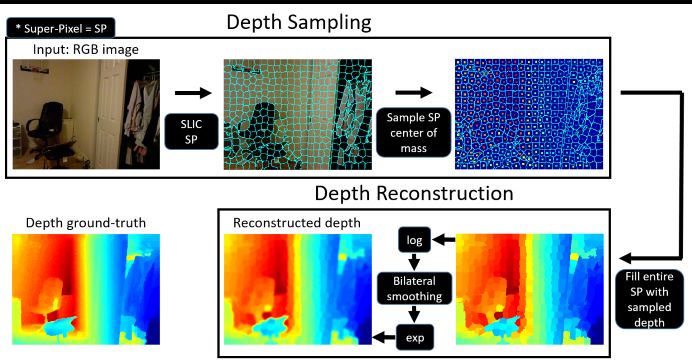}
    \caption{Algorithm block diagram.}
    \label{fig:algorithm}
\end{figure*}

\subsection{Algorithm design requirements}
Several requirements are vital to the design of such an algorithm: It needs to capture well the shape and boundaries of objects, to be computationally fast and memory efficient, and to have control on the number of samples. Surprisingly, all these requirements coincide with those for super-pixels (SPs) \cite{achanta2012slic} - an over-segmentation technique applied to RGB images. This led us to the following algorithm.

\subsection{Proposed algorithm}
The proposed algorithm is divided into two parts, sampling and reconstruction. It includes the following steps:
\begin{itemize}
    \item {\bf Sampling:}
    \begin{enumerate}[label=S.\arabic*]
        \item A super-pixel map is generated from the RGB image using SLIC \cite{achanta2012slic}. The desired number of SPs is set to $n$. The SPs compactness is adjusted to high value to ensure regularly shaped SPs.
        \item \label{s2} For each SP, the SP center of mass (CoM) is computed by calculating the mean of the $(x,y)$ coordinates of all pixels in the SP.
        A depth sample is taken at the CoM location. If the CoM is located outside of the SP (for some non-convex SP), the depth sample is taken at the closest location to the CoM of the SP.
    \end{enumerate}
    \item {\bf Reconstruction:}
    Our reconstruction is based on the samples and SPs of the sampling stage.
    \begin{enumerate}[label=R.\arabic*]
        \item For each SP, a single depth measurement is available, thus a zero-order estimation is performed. That is, the entire SP takes the depth value of the sample. Let $d_0$ be the resulting depth image.
        \item $d_{log}$ is calculated by $d_{log}=\log{(d_0+1)}$.
        \item A bilateral filter \cite{tomasi1998bilateral} is applied over $d_{log}$. The filter's parameters are fixed for a given number of samples $n$ and type of scene (road / room). Let $d_{BF}$ be the bilateral filter result.
        \item The final dense reconstructed depth image, $d_r$, is calculated by $d_r=\exp{(d_{BF})}-1$.
        \end{enumerate}
\end{itemize}

See Fig. \ref{fig:algorithm} for a high-level diagram of our framework.

\subsection{Principles of the algorithm}

\graphicspath{{./figures/algorithm/ablation_study/}}
% Ablation study
\begin{figure}[htb]
    \centering
    \includegraphics[width=0.75\textwidth]{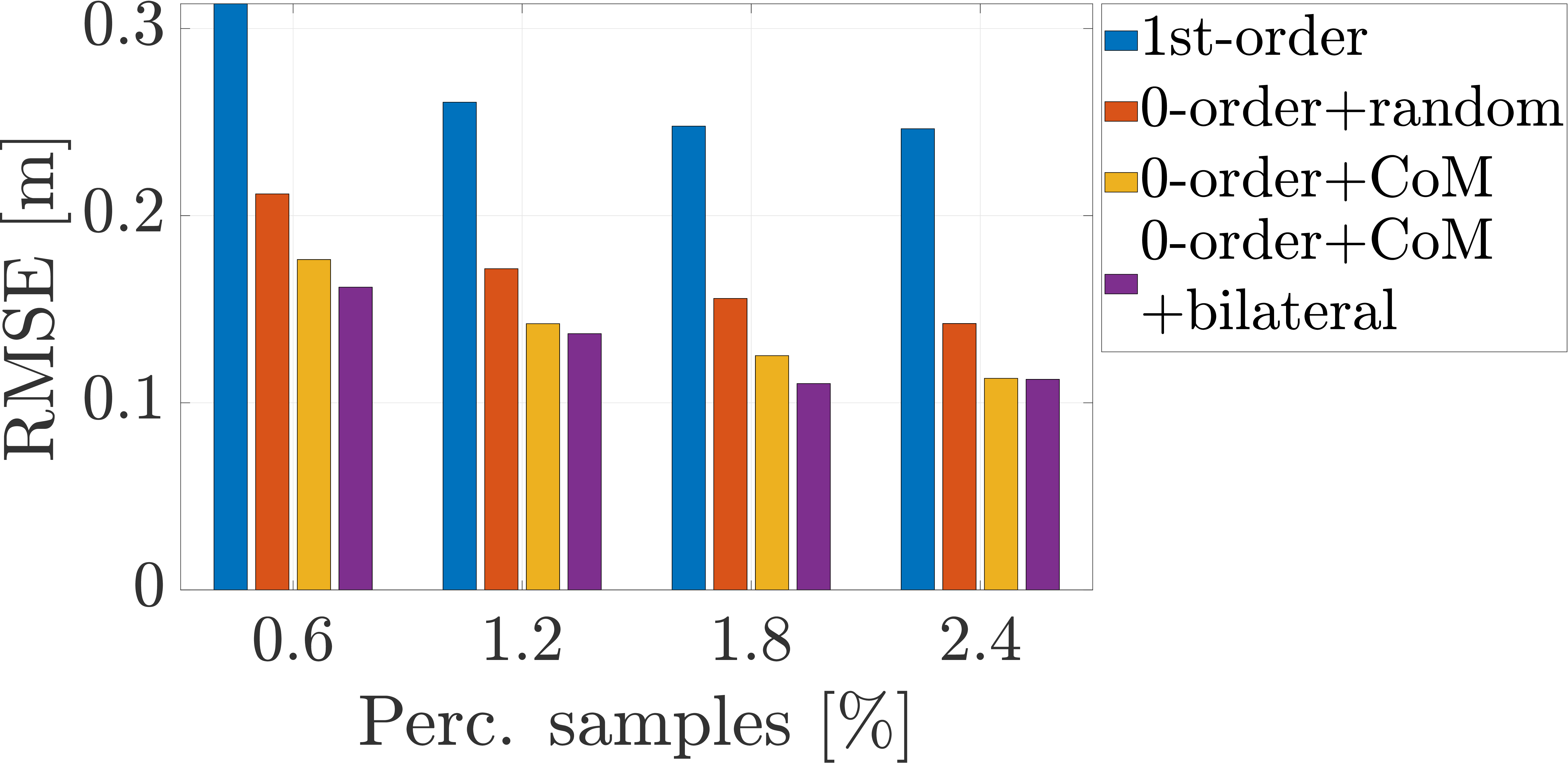}
    \caption{Reconstruction variants study on NYU-Depth-v2 dataset. We compare 1st-order with 3 samples per segment to 3 variants of 0-order reconstruction, all with the same number of samples.}
    \label{fig:ablation_study}
\end{figure}

% 0-order reconstruction is better than 1st-order regardless of sampling percentage. Sampling at SP center of mass is preferred than sampling a random point inside the SP. Applying bilateral filter over the reconstructed depth improves the result for low sampling rates.

\subsubsection{Sampling}
\textbf{Sampling based on RGB segmentation.} This follows  the model and relations between RGB edges and depth discontinuities, discussed in previous section.
%using the relatively high values of $P_{rgb-d}$. 
    
\noindent \textbf{Sampling at center of mass of segment.} There are several reasons for this choice: it reduces an inherent uncertainty near discontinuities. Secondly, for piece-wise planar depth regions, the center of mass minimizes the RMSE. Moreover, practical depth sensing technologies have a finite spatial resolution and cannot sample well near depth discontinuities.
Fig. \ref{fig:ablation_study} demonstrates that sampling at SP CoM leads to a more accurate depth reconstruction results than sampling a random pixel location inside the SP.

\begin{figure}[htb]
    \centering
    \begin{subfigure}{0.23\textwidth}
        \centering
        \frame{\includegraphics[height=1.0in]{rgb.png}}
        \caption{RGB}
    \end{subfigure}
    \begin{subfigure}{0.23\textwidth}
        \centering
        \includegraphics[height=1.0in]{depth.png} 
    \caption{Depth GT}
    \end{subfigure}
    \begin{subfigure}{0.23\textwidth}
        \centering
        \includegraphics[height=1.0in]{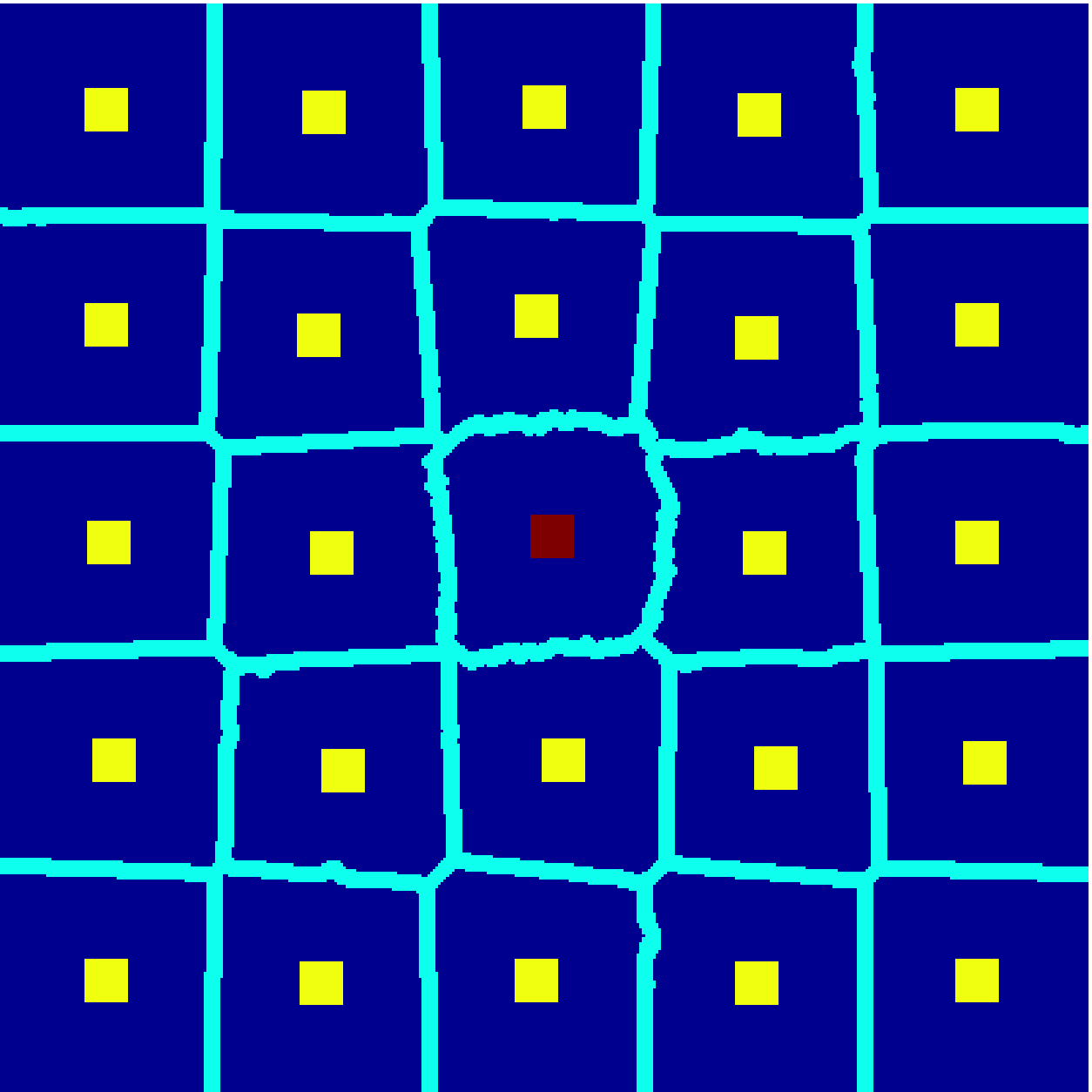} 
    \caption{Samples}
    \end{subfigure}
    \begin{subfigure}{0.23\textwidth}
        \centering
        \includegraphics[height=1.0in]{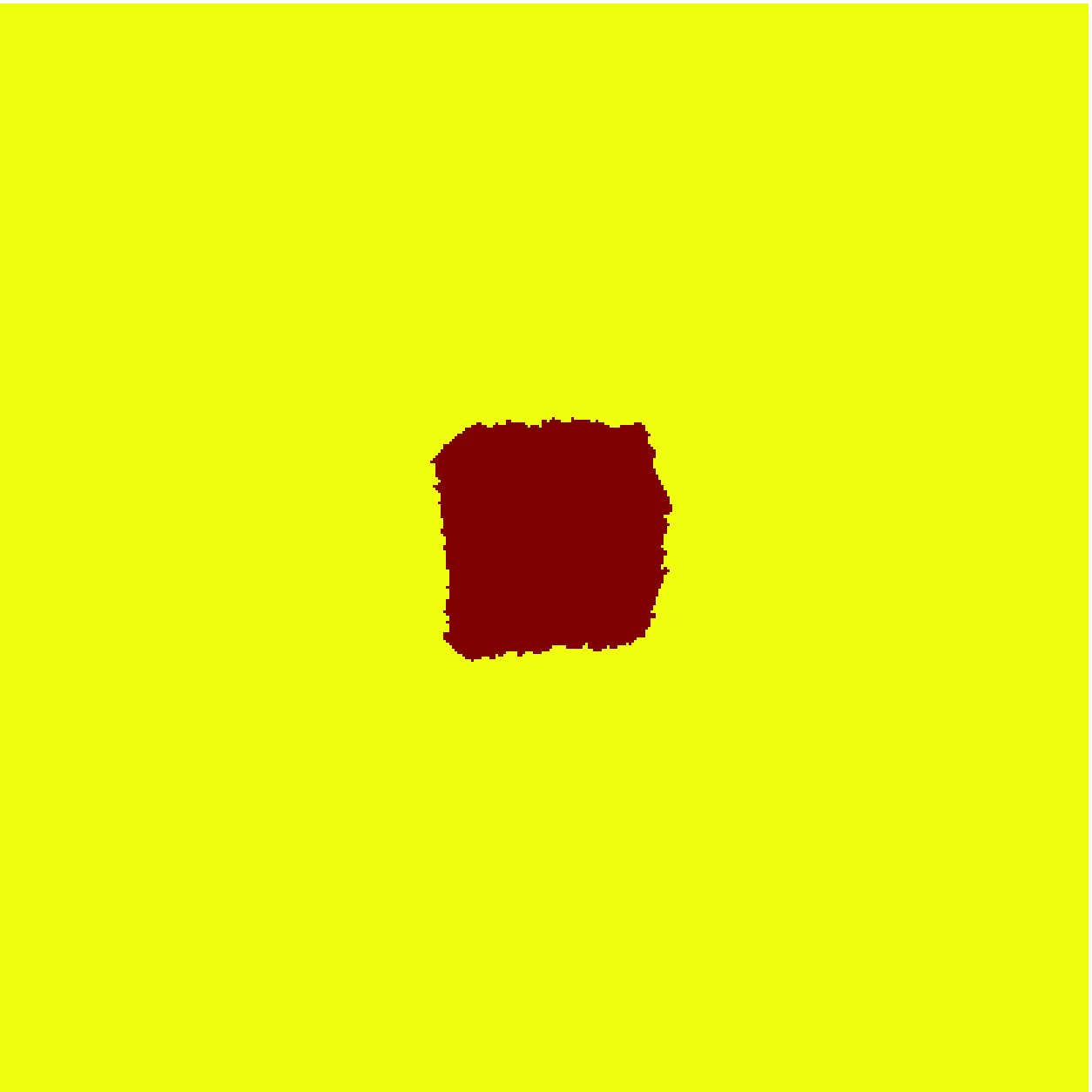} 
    \caption{Recon.}
    \end{subfigure}
    
% \graphicspath{{./figures/algorithm/toy_example_2/narrow/}}
% \begin{figure}[t!]
%     \centering
%     \begin{subfigure}{0.20\textwidth}
%         \centering
%         \includegraphics[height=0.5in]{rgb.eps} 
%         \caption{RGB}
%     \end{subfigure}
%     \begin{subfigure}{0.20\textwidth}
%         \centering
%         \includegraphics[height=0.5in]{depth.eps} 
%     \caption{Depth GT}
%     \end{subfigure}
%     \\
%     \begin{subfigure}{0.20\textwidth}
%         \centering
%         \includegraphics[height=0.5in]{sp_samp.eps} 
%     \caption{Sampling pattern}
%     \end{subfigure}
%     \begin{subfigure}{0.20\textwidth}
%         \centering
%         \includegraphics[height=0.5in]{sp_recon.eps} 
%     \caption{Reconstruction}
%     \end{subfigure}
    \caption{Toy example 2: Camouflaged object. In some rare cases, the object is camouflaged and can hardly be detected in the RGB image. In this case image-guided sampling fails. As our method is based on super-pixels, when there are no distinct edges, the method degenerates in a natural manner to classical grid sampling.}
    \label{fig:algorithm_toy2}
\end{figure}

\noindent \textbf{Why super-pixels.} Sampling with super-pixels enables measuring small elements. It also limits reconstruction error since the size of the segment is limited. When the depth discontinuity is not well reflected in the RGB (we term it \emph{camouflaged objects}) the sampling reduces to an approximate grid-sampling scheme, which provides a lower bound on the resolution. This is illustrated in Fig. \ref{fig:algorithm_toy2}.

%In addition, the decision of setting high SP compactness was made to obtain regularly shaped zero-order reconstructed depth.

\subsubsection{Reconstruction}
\begin{figure}[htb]
    \centering
    \begin{subfigure}{0.3\textwidth}
        \centering
        \includegraphics[height=1.1in]{rgb.png} 
    \caption{RGB}
    \end{subfigure}
    \begin{subfigure}{0.3\textwidth}
        \centering
        \includegraphics[height=1.1in]{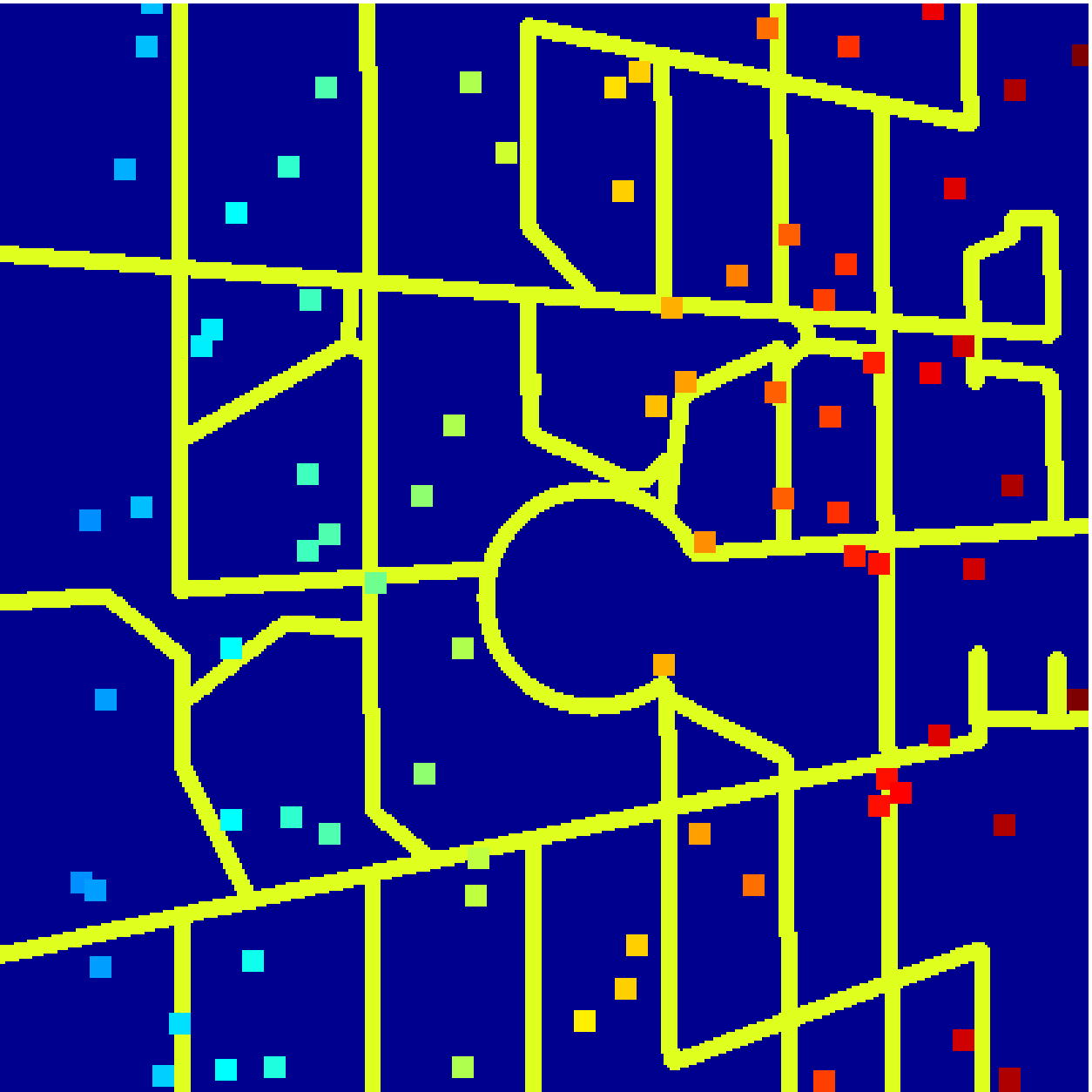}
    \caption{Large seg. samp.}
    \end{subfigure}
    \begin{subfigure}{0.3\textwidth}
        \centering
        \includegraphics[height=1.1in]{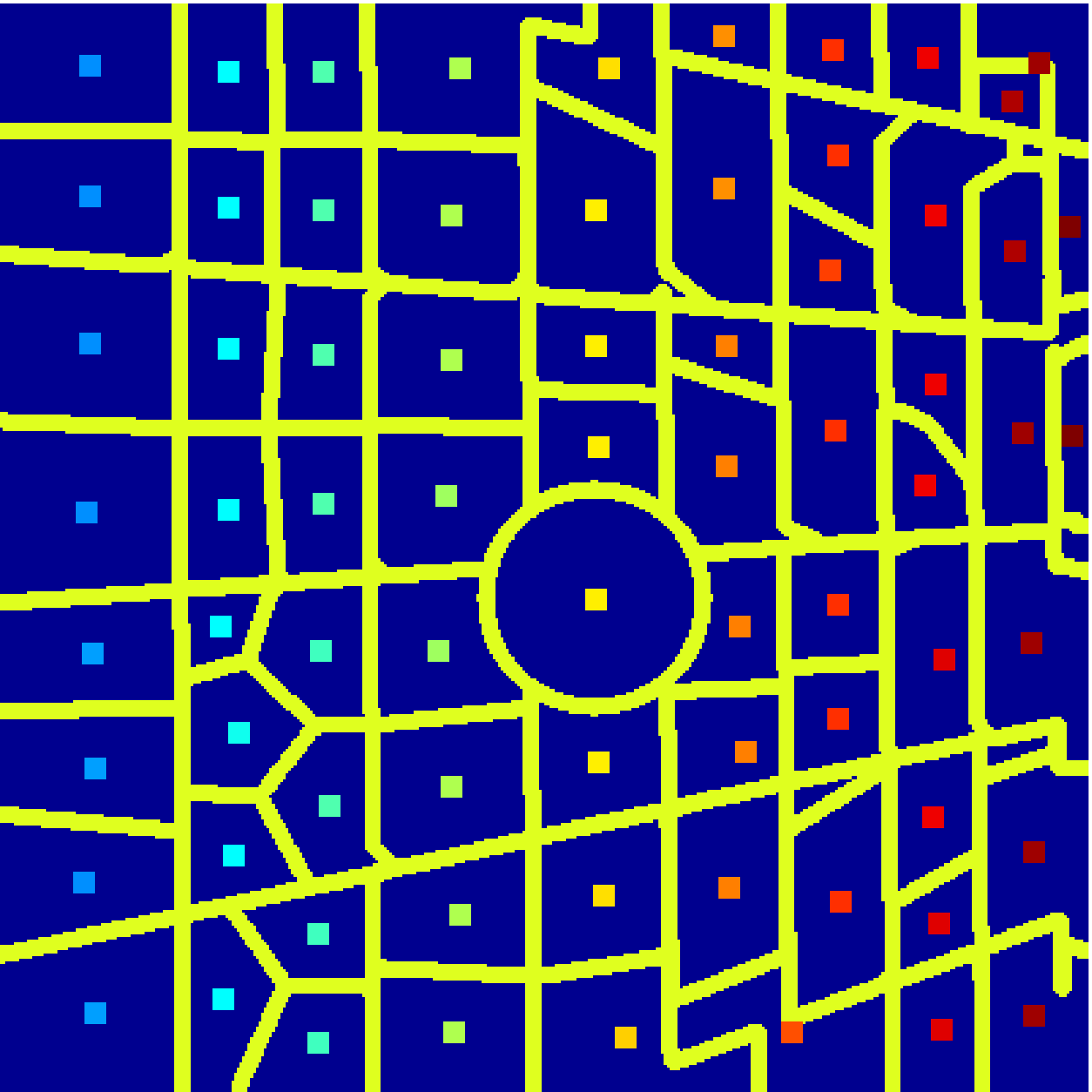} 
    \caption{Small seg. samp.}
    \end{subfigure}
    \\
    \begin{subfigure}{0.3\textwidth}
        \centering
        \includegraphics[height=1.1in]{depth.png} 
    \caption{Depth GT}
    \end{subfigure}
    \begin{subfigure}{0.3\textwidth}
        \centering
        \includegraphics[height=1.1in]{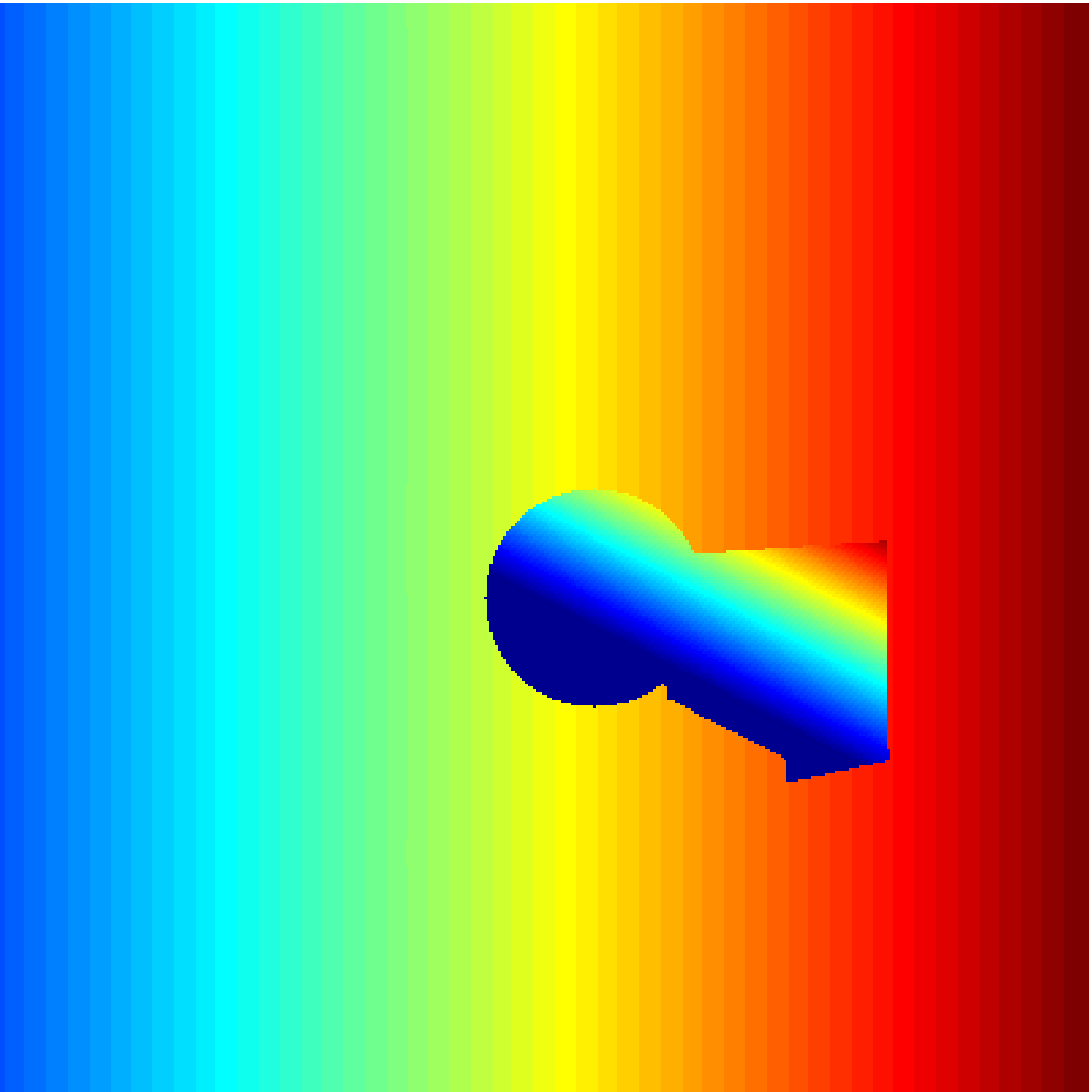} 
    \caption{1st-order recon.}
    \end{subfigure}
    \begin{subfigure}{0.3\textwidth}
        \centering
        \includegraphics[height=1.1in]{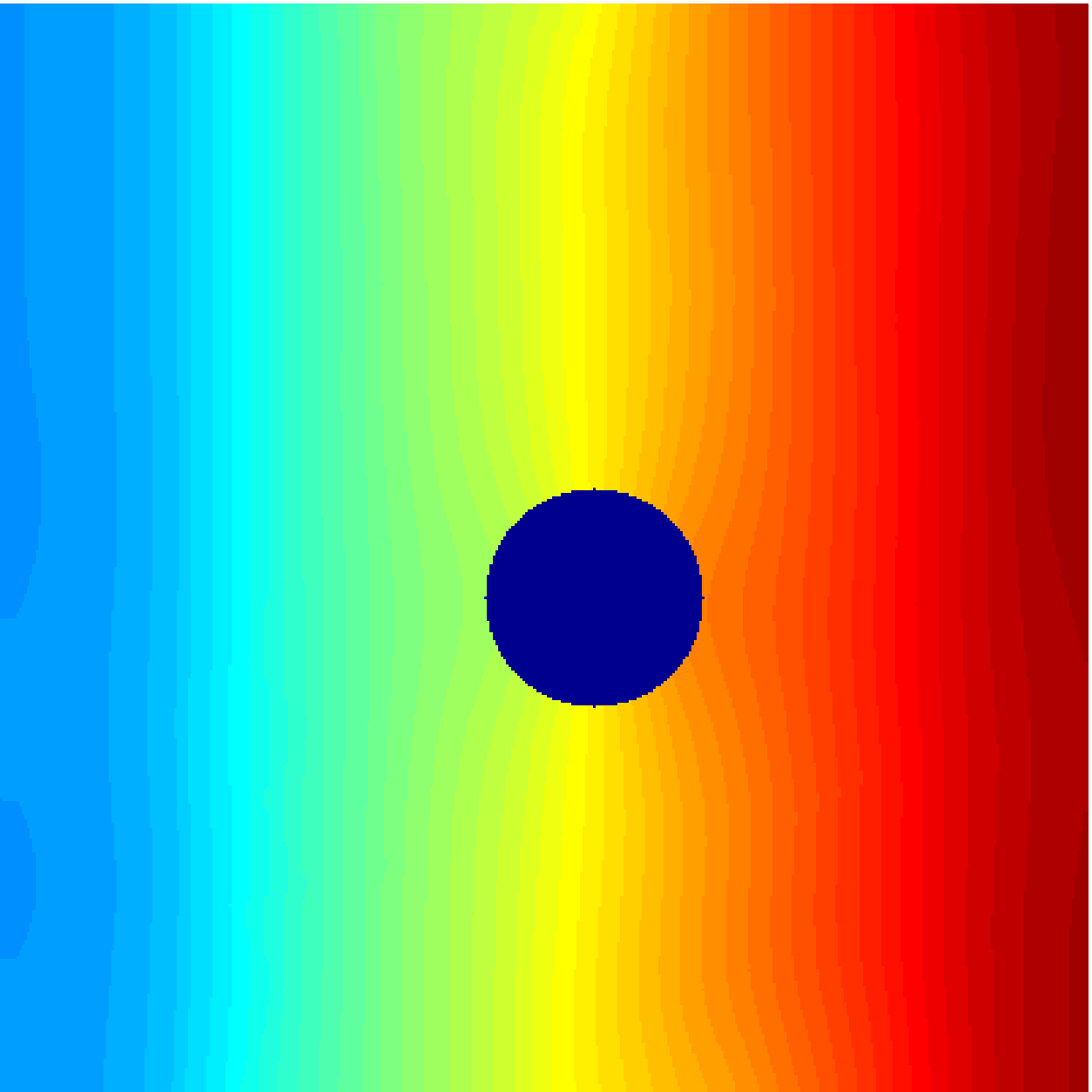} 
    \caption{Our recon.}
    \end{subfigure}
    \caption{Toy example 1: Comparing 1st-order estimation of larger segments (middle row, RMSE=55.0) to zero-order estimation for smaller segments, following nonlinear smoothing (bottom, RMSE=7.7). In both cases 75 samples were used. The latter approach allows for smaller objects to be well reconstructed at the expense of slight staircasing artifacts in larger flat regions (roads, walls etc.). }
    %As our primary objective is to recover well small objects - we adopted the latter sampling and reconstruction strategy.}
    \label{fig:algorithm_toy1}
\end{figure}

\textbf{0-order vs. 2D linear reconstruction in each segment.} At a first glance, it seems natural to estimate by SPs the subdomains $\Omega_i$ of the model, which require 3 samples to obtain a plane approximation in the region of the SP. However, we found out that this is not an optimal strategy.
A better approach is to increase the number of segments by a factor of 3 and to sample once each segment. This allows to increase the overall resolution (or \emph{smallest object size}) of the system while still being able to recover reasonably well large planar segments, with a proper nonlinear filtering operation (see below). The depth of the smallest objects that the system can measure are estimated by a constant value. This facilitates the detection of poles, signs and small obstacles at a low sampling cost.
Fig. \ref{fig:ablation_study} demonstrates that 0-order reconstruction is much more accurate than linear reconstruction in terms of RMSE, for a given sampling budget $n$. In  Fig. \ref{fig:algorithm_toy1} the ability to detect well small objects by 0-order estimation is illustrated.

\noindent \textbf{Bilateral filtering.} 
Having more SPs with zero-order estimation allows to sample well small objects. However, now large flat regions are heavily degraded by staircasing artifacts. Our proposed solution is to apply a fast, nonlinear, edge preserving filter \cite{tomasi1998bilateral}. It is designed such that actual depth discontinuities are preserved, whereas false edges, which stem from the 0-order estimation, are smoothed out. Due to the log function, smoothing is relative to the depth. This approximates well the piece-wise planar model for large regions, such as walls and roads, yielding also lower RMSE, as seen in Fig. \ref{fig:ablation_study}.
As can be seen in Fig. \ref{fig:algorithm_toy1}, the artifacts in the reconstruction of the large planar background region are quite minimal.

\graphicspath{{./figures/algorithm/toy_example_1/}}

% Toy example 2
\graphicspath{{./figures/algorithm/toy_example_2/}}

% \subsection{Semantic Data Usage}
% For applications such as autonomous car driving, we can use semantic knowledge obtained from the RGB image in order to avoid sampling some areas, or dramatically decrease the number of samples there. 

\subsubsection{MTF analysis}
We aim to measure the spatial resolution of our sampling and reconstruction strategy. We use modulation transfer function (MTF) - a standard tool for characterizing the resolution of imaging systems \cite{boreman2001modulation}. Our chart is based on the Siemens Star testchart modified for RGB-guided depth sampling.  
%\textcolor{red}{mention that the original is black and white?} 
%and measured . For RGB guidance \ref{mtf_rgb}, 
We modified the original testchart by a road scene, replacing the white regions with typical background content and the black regions with typical foreground content. Depth is assumed to take binary values, as in the original chart. 
The MTF calculation is explained in detail in \cite{loebich2007digital}.
% we give a short summary in the supplementary material. 

\begin{figure}[htb]
    \centering
    \includegraphics[width=0.85\textwidth]{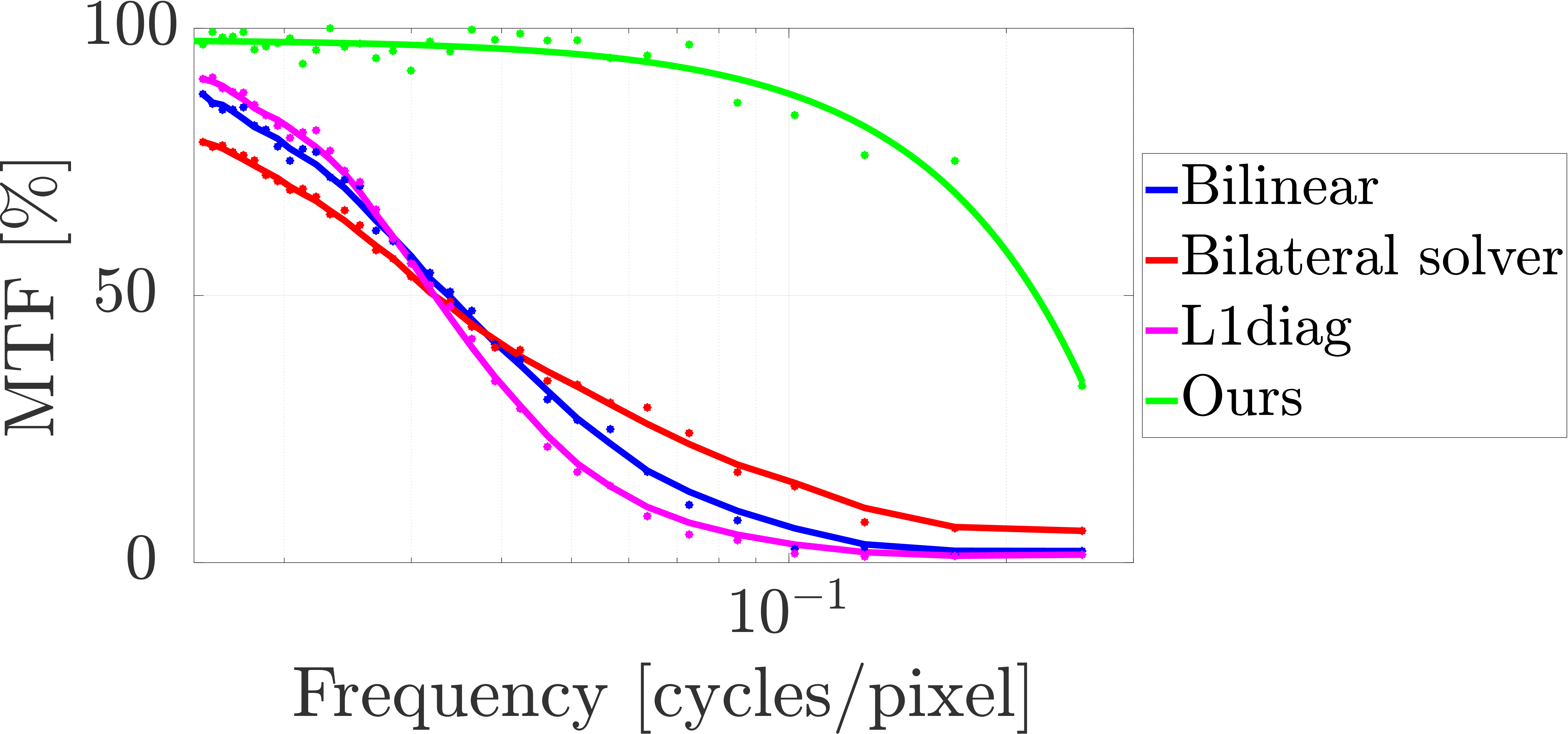}
    \caption[Caption for LOF]{MTF comparison between bilinear\protect\footnotemark, bilateral solver \cite{barron2016fast}, L1diag \cite{ma2017sparse} and ours. Our method achieves significantly higher resolution at the same sampling rate.}
    \label{fig:mtf_quant}
\end{figure}

\begin{figure}[htb]
    \centering
    \begin{subfigure}{0.3\textwidth}
        \centering
        \includegraphics[height=1.3in]{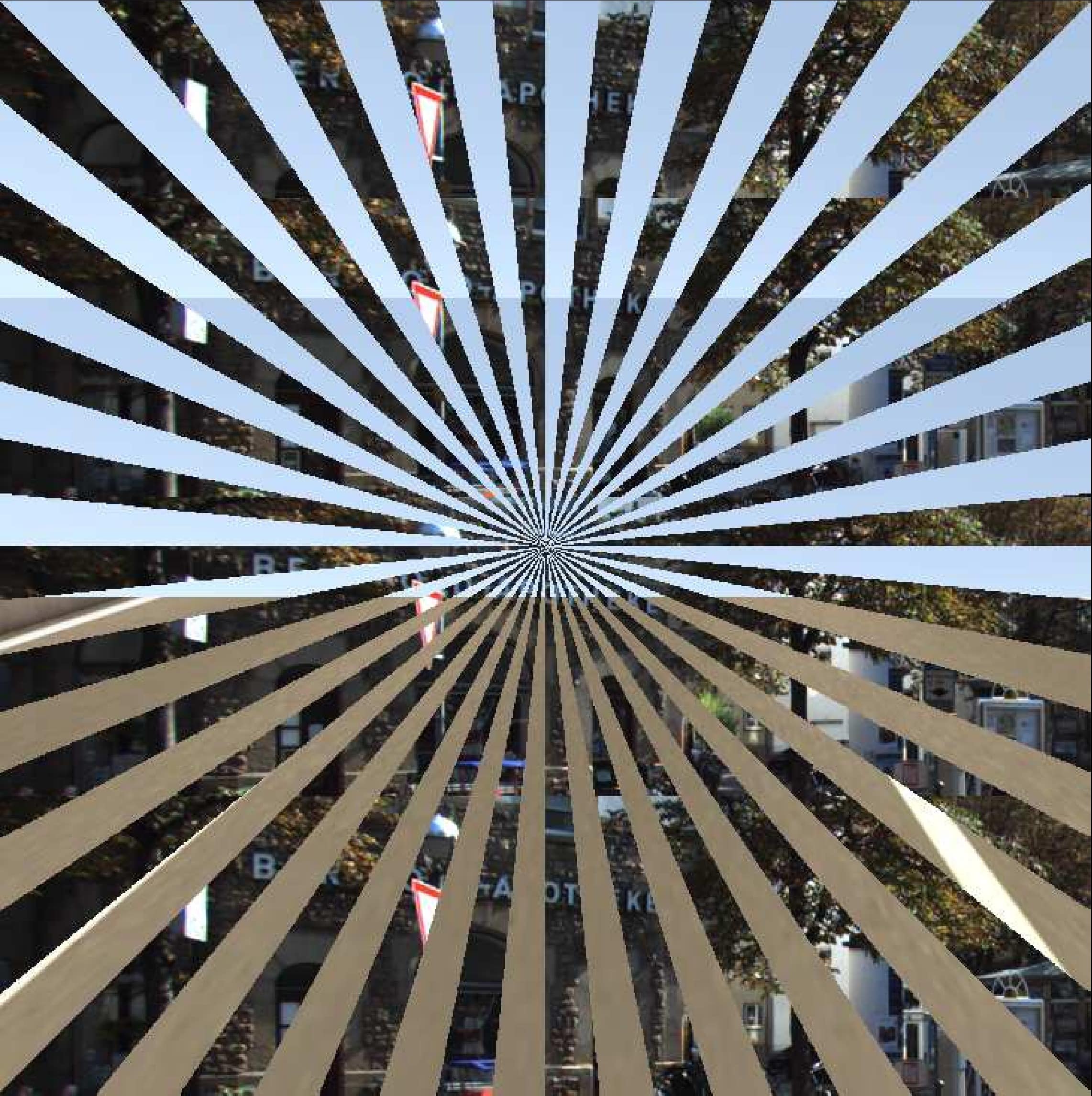} 
        \caption{RGB}
        \label{mtf_rgb}
    \end{subfigure}
    \begin{subfigure}{0.3\textwidth}
        \centering
        \includegraphics[height=1.3in]{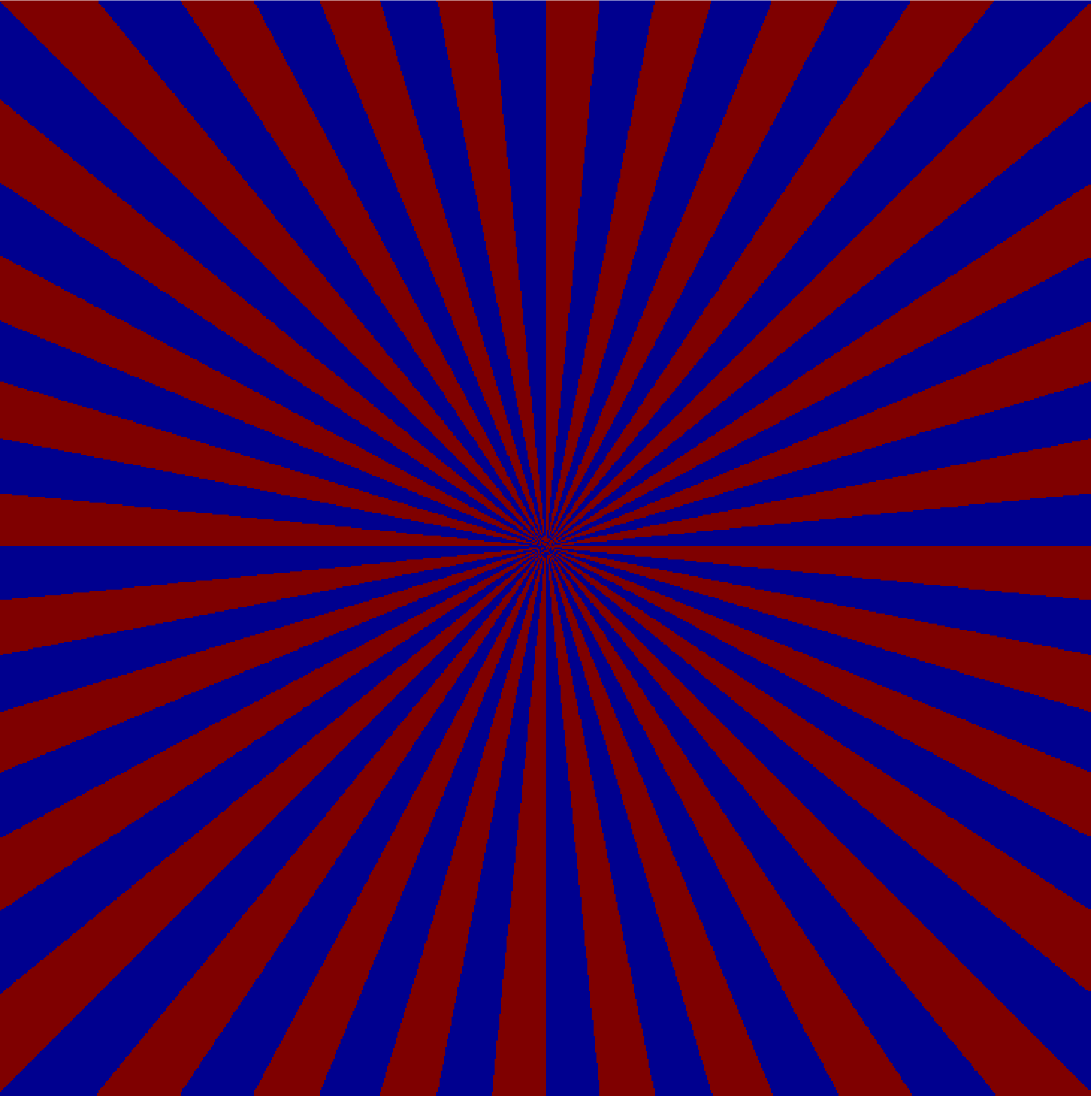} 
        \caption{Depth GT}
        \label{mtf_depth}
    \end{subfigure}
    \begin{subfigure}{0.3\textwidth}
        \centering
        \includegraphics[height=1.3in]{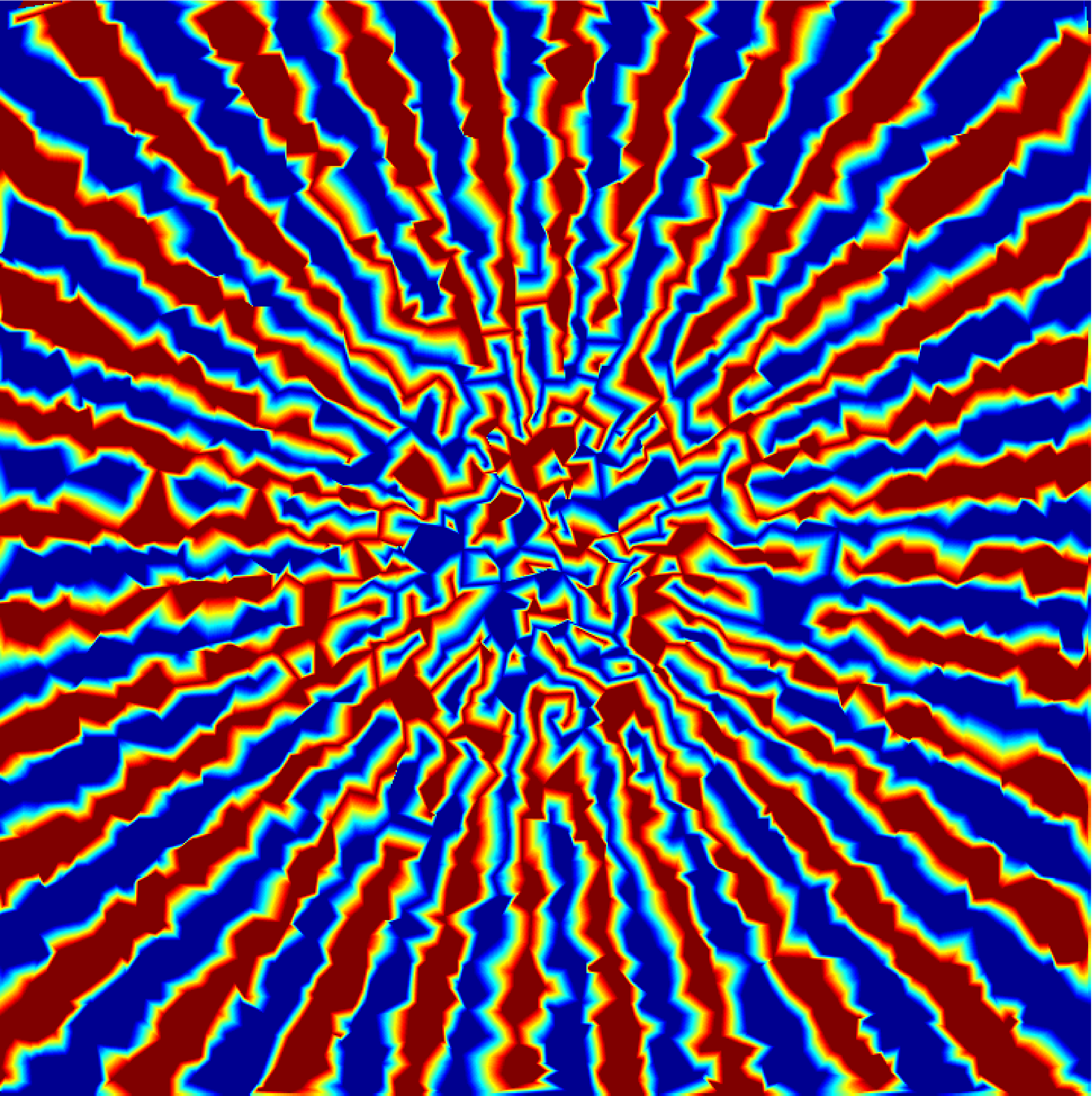} 
        \caption{Bilinear}
    \end{subfigure}
    \\
    \begin{subfigure}{0.3\textwidth}
        \centering
        \includegraphics[height=1.3in]{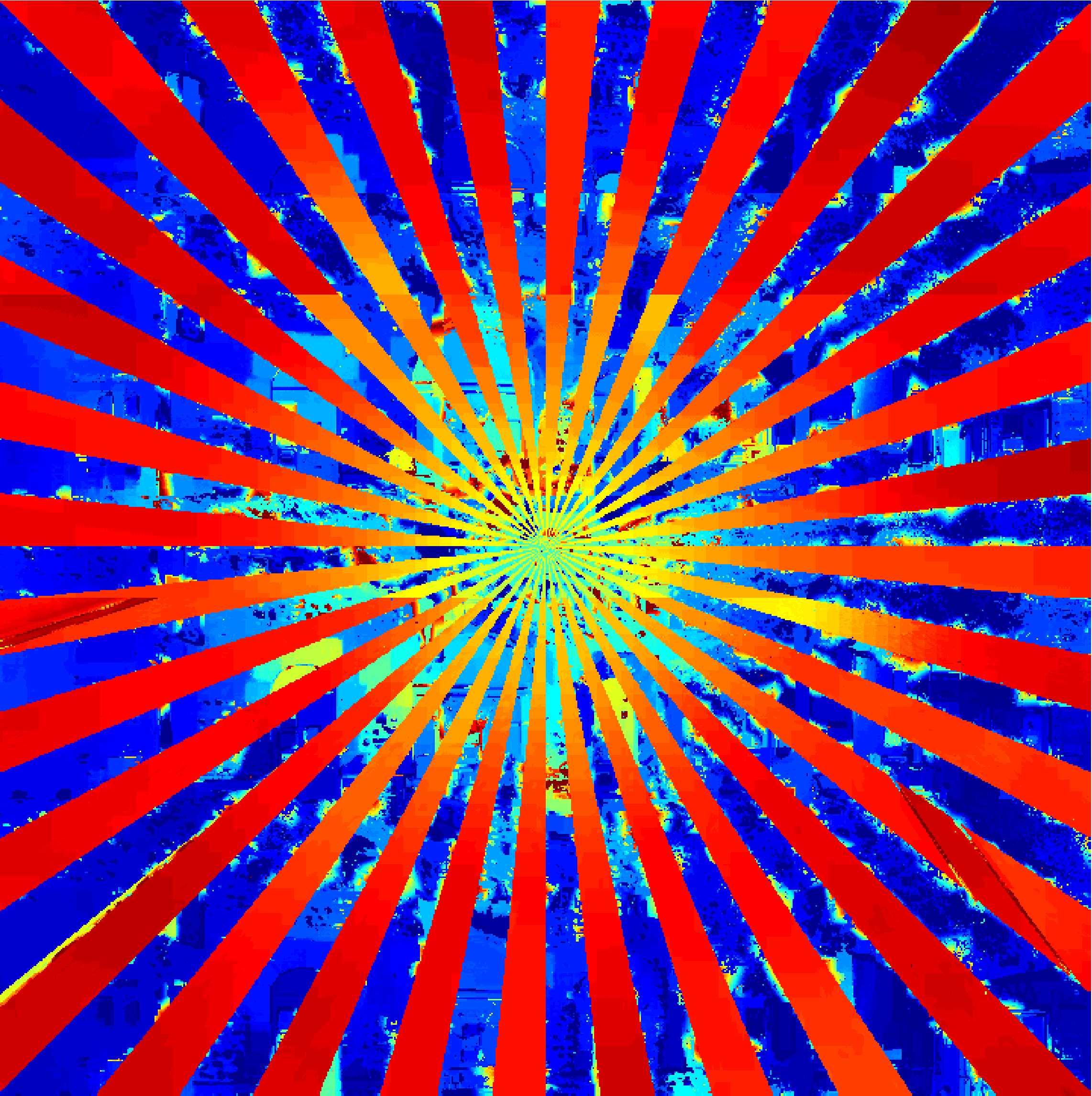}
        \caption{Bilat. solver \cite{barron2016fast}}
    \end{subfigure}
    \begin{subfigure}{0.3\textwidth}
        \centering
        \includegraphics[height=1.3in]{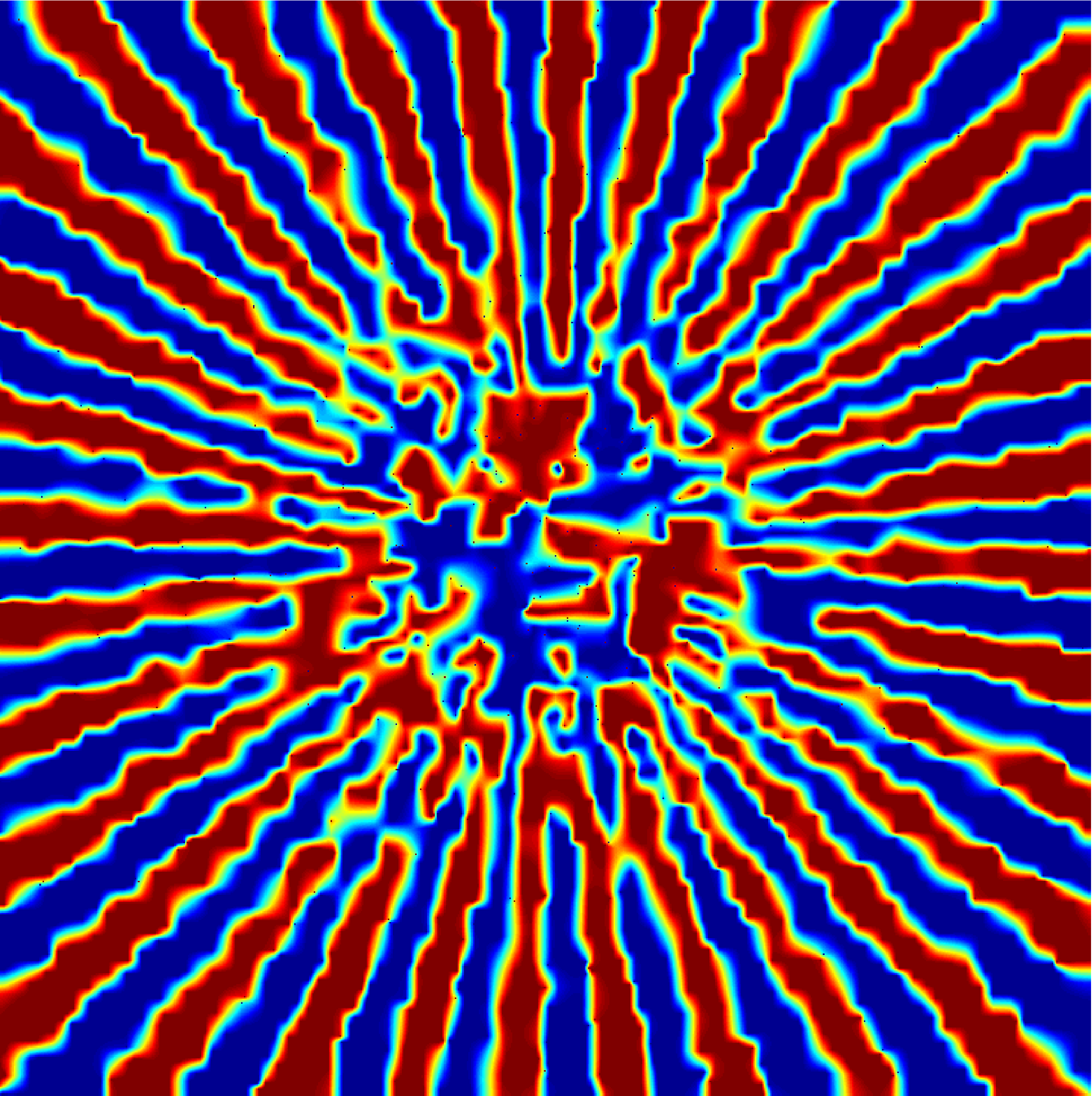} 
        \caption{L1diag \cite{ma2017sparse}}
    \end{subfigure}
    \begin{subfigure}{0.3\textwidth}
        \centering
        \includegraphics[height=1.3in]{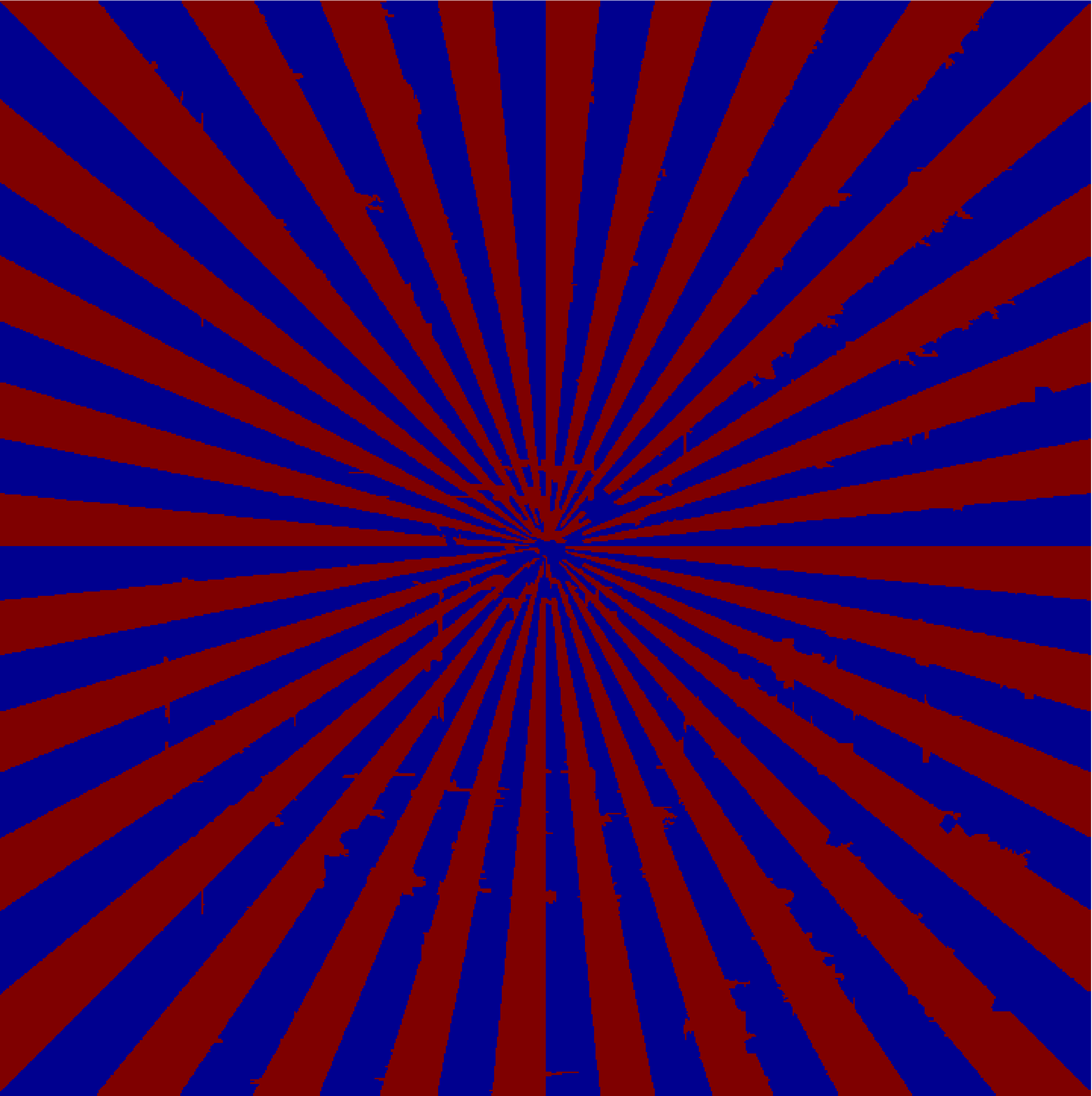} 
        \caption{Ours}
    \end{subfigure}
    \caption{Qualitative results for depth reconstruction on our modified MTF road-like testchart with $n=5000$ samples (1\% ratio).}
    \label{fig:mtf_qual}
\end{figure}

Fig. \ref{fig:mtf_quant} presents the results computed from the reconstructed images in Fig. \ref{fig:mtf_qual}. One can observe the clear increase of resolution of the proposed method, compared to RGB-guided and non-guided depth completion approaches.

% MTF - quantitative
\graphicspath{{./figures/algorithm/MTF/}}

\footnotetext{\label{bilinear}We use Delaunay triangulation \cite{amidror2002scattered} to perform a bivariate linear interpolation.}

% MTF - qualitative
\graphicspath{{./figures/algorithm/MTF/}}

%-------------------------------------------------------------------------

\section{Experiments}
We evaluate the performance of our depth sampling and reconstruction method and compare it to other approaches. For all other examined methods we simulate uniform random depth samples at different sparsity levels. We define pixel density as the ratio between the number of sampled pixels to the total number of pixels in the image. 
To demonstrate the generalization of our algorithm, we use two distinct datasets for evaluation - one for outdoor scenarios and one for indoor scenarios. For the outdoor dataset we also evaluate over a subset of image areas focusing on small obstacles in the image. Later, we make a qualitative comparison between different sampling patterns and show that ours leads to better result. Finally, we show initial experimental results, using a real system we built, based on the proposed principles.
We measure performance on all experiments with RMSE (root mean squared error), and also report the REL (relative absolute error) metric on NYU-Depth-v2.
% In this section, we are led by two main objectives. First, we intent to show that our sampling pattern is preferred over others in any case. Second, we anticipate better results from our full sampling and reconstruction compared to other depth completion methods. Regarding the first objective, we compare different sampling strategies by performing separate depth completion methods over the sampled values. Concerning the second objective, we add the output of our complete algorithm. We repeat these processes with various sampling rates. Moreover, to demonstrate the generalization of our algorithm, we use several datasets distinct from each other for evaluation.

\subsection{Outdoor data (Synthia)}
The Synthia dataset \cite{ros2016synthia} provides synthetic RGB, depth and semantic images for urban driving scenarios. We use synthetic data as for now there is no large non-synthetic dataset that provides a dense and accurate depth map. We need a dense depth map to be able to sample at any given point. Accuracy is required especially to show the increased resolution we obtain. Thus, two large real-life datasets do not apply: KITTI depth completion benchmark \cite{uhrig2017sparsity} has semi-dense depth, and Cityscapes \cite{cordts2016cityscapes} has low resolution depth, which is in some cases inaccurate. More technical details are given in Section \ref{model_section}. We make the evaluation on 0-100m depth range, which is similar to the range of a typical vehicle-mounted LiDAR. 

\begin{figure}[htb]
    \centering
    \begin{subfigure}{0.8\textwidth}
        \centering
        \includegraphics[height=0.25in]{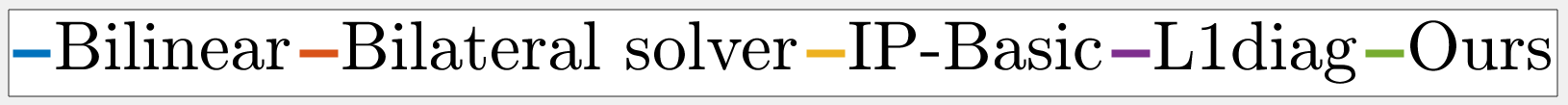}
    \end{subfigure}
    \\
    \begin{subfigure}{0.45\textwidth}
        \centering
        \includegraphics[height=1.35in]{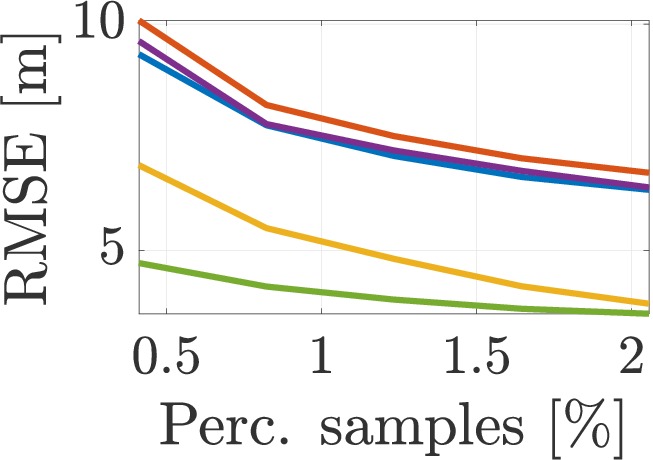}
        \caption{Synthia}
        \label{fig:synthia_quant}
    \end{subfigure}
    \begin{subfigure}{0.45\textwidth}
        \centering
        \includegraphics[height=1.35in]{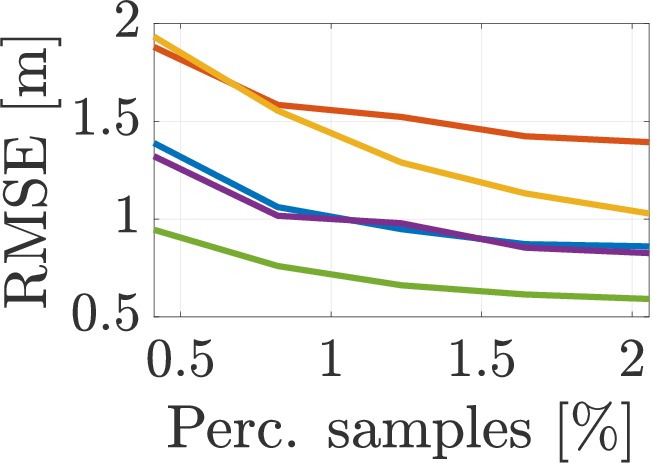}
        \caption{Synthia obstacles}
        \label{fig:synthia_obst_quant}
    \end{subfigure}   
    \caption{Quantitative comparison on Synthia (left) and obstacles (right) datasets between bilinear interpolation, bilateral solver \cite{barron2016fast}, IP-Basic \cite{ku2018defense}, L1diag \cite{ma2017sparse} and ours.}
\end{figure}

\begin{figure*}[htb]
    \centering
    \includegraphics[width=0.95\linewidth]{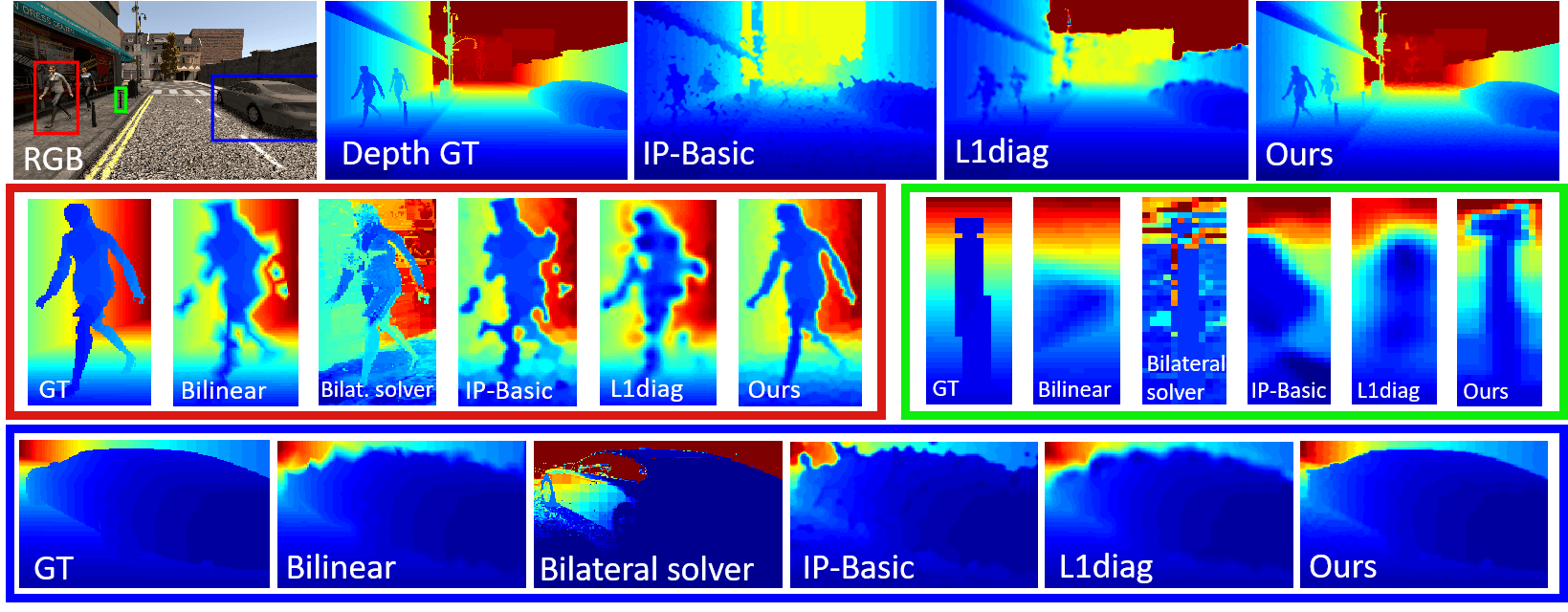}
    \caption{Qualitative results of depth completion on Synthia and our obstacles dataset at 2\% density (5000 samples). Tested completion methods: bilinear interpolation, bilateral solver \cite{barron2016fast}, IP-Basic \cite{ku2018defense}, L1diag \cite{ma2017sparse} and ours.}
    \label{fig:synthia_qual}
\end{figure*}

{\bf Full scene experiment.}
Quantitative results are presented in Fig. \ref{fig:synthia_quant}. We achieve a 30\% lower RMSE than the second best at 0.45\% density, and keep having the best result through all densities. We also evaluated a deep-learning method \cite{eldesokey2018propagating} trained on KITTI benchmark \cite{uhrig2017sparsity}, but it failed to obtain any comparable results.
Qualitative comparison is shown in Fig. \ref{fig:synthia_qual}, exhibiting precise and sharp reconstruction, especially of small objects. 
%\textcolor{red}{the optimal RMSE is not in this range}

\graphicspath{{./figures/experiments/}}
% synthia - quantitative

% synthia - qualitative

{\bf Obstacles set.}
To enable evaluation over important objects in the image and to reduce the impact of far background on the performance measurement, we derived from Synthia a set of 100 obstacles, which we refer to as \emph{the obstacles dataset}. 
% A demonstration of the dataset is provided in the supplementary material.
%
We applied sampling and reconstruction over the entire image, but now evaluate over the obstacle mask. Quantitative results are presented in Fig. \ref{fig:synthia_obst_quant}, and qualitative comparison is shown in Fig. \ref{fig:synthia_qual}. Table \ref{tab:synthia_obst_quant_swap} compares the number of samples $n$ required to achieve certain levels of accuracy. \emph{We require 3-4 times less samples for a given RMSE.}

% % obstacles - examples
% \begin{figure}[htb]
%     \begin{subfigure}{0.45\textwidth}
%         \centering
%         \includegraphics[width=0.85\textwidth]{synthia_obstacles_dataset.png} 
%         \caption{Synthia}
%         \label{fig:synthia_obst}
%     \end{subfigure}
%     \\
%     \begin{subfigure}{0.45\textwidth}
%         \centering
%         \includegraphics[width=0.85\textwidth]{nyuv2_obstacles_dataset.png} 
%         \caption{NYU-Depth-v2}
%         \label{fig:nyuv2_obst}
%     \end{subfigure}
%     \caption{Examples from our obstacles datasets.}
%     \label{fig:obst_example}
% \end{figure}

% synthia obstacles - samples per RMSE
\begin{table}[htb]
    \centering
    \begin{tabular}{c|c|c|c}
        RMSE [m] & 0.6 & 0.75 & 0.95 \\
        \hline
        Bilinear interp. & 7.87\% & 2.90\% & 1.23\% \\
        Bilateral solver \cite{barron2016fast} & $>$ 8\% & $>$ 8\% & $>$ 8\% \\
        IP-Basic \cite{ku2018defense} & $>$ 8\% & 4.78\% & 2.42\% \\
        L1diag \cite{ma2017sparse} & 7.09\% & 3.08\% & 1.33\% \\
        Ours & \textbf{1.89\%} & \textbf{0.86\%} & \textbf{0.40\%} \\
    \end{tabular}
    \caption{Quantitative comparison for depth completion of required sampling percentage per RMSE on our Synthia obstacles dataset. Our method is 3-4 times more economic than second best.}
    \label{tab:synthia_obst_quant_swap}
\end{table}

\begin{figure*}[htb]
    \centering
    \includegraphics[width=0.95\linewidth,height=1.5in]{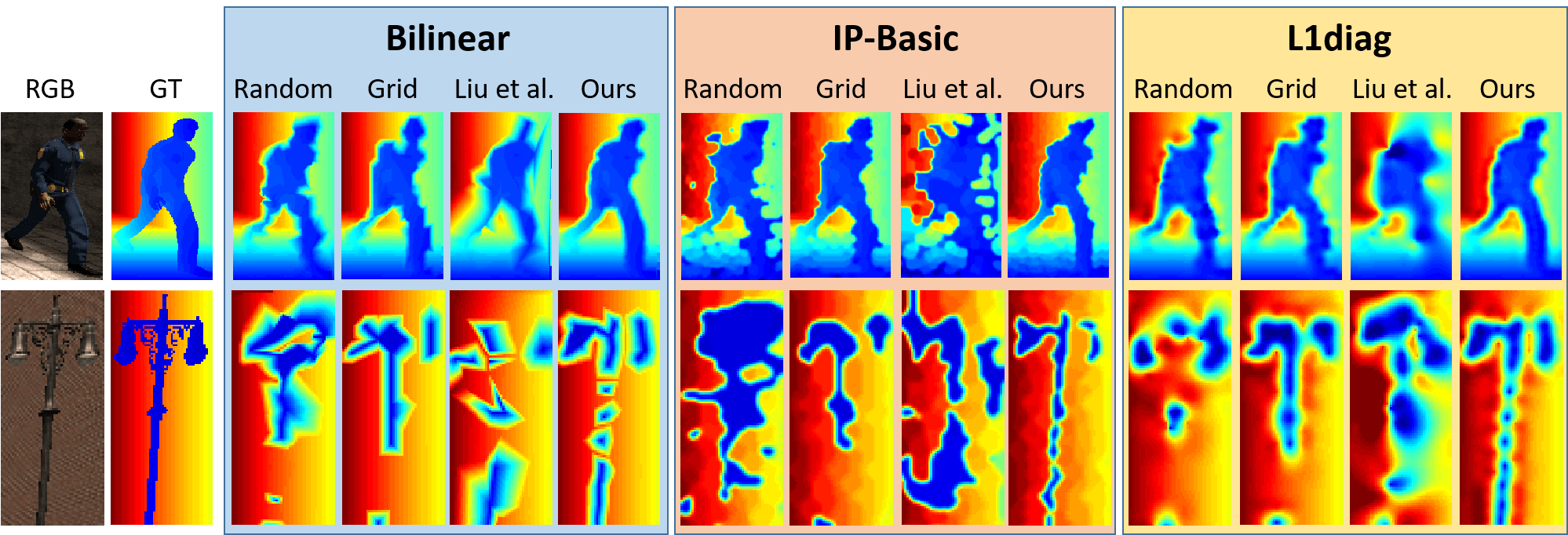}
    \caption{\textbf{Sampling methods comparison:} 4 sampling patterns are compared for different reconstruction method: uniform random, grid, Liu et al. \cite{liu2015depth} and ours (sampling only). The reconstructions are bilinear interp., IP-Basic \cite{ku2018defense} and L1diag \cite{ma2017sparse}}
    \label{fig:sampling_comparison}
\end{figure*}

{\bf Sampling only.}
We claim that using only our proposed sampling pattern, any completion method can achieve better results than using other existing sampling patterns, especially for small objects in the scene. Fig. \ref{fig:sampling_comparison} proves this qualitatively for 3 distinct reconstruction methods.

% synthia - sampling comparison

% NYUv2 - quantitative
\begin{table}[htb]
    \centering
    \begin{tabular}{c|c|c|c}
        Samples & Method & RMSE [m] & REL \\
        \hline
        200 & Bilinear interp. & 0.257 & 0.047\\
        200 & L1diag \cite{ma2017sparse} & 0.236 & 0.044 \\
        225 & Liao et al. \cite{liao2017parse} & 0.442 & 0.104\\
        200 & Ma et al. \cite{ma2018sparse} & 0.230 & 0.044\\
        200 & HMS-Net \cite{huang2018hms} & 0.233 & 0.044\\
        200 & Li et al. \cite{li2018depth} & 0.256 & 0.046\\
        200 & Ours & \textbf{0.211} & \textbf{0.035}\\
    \end{tabular}
    \caption{Quantitative comparison with state-of-the-art on the NYU-Depth-v2 dataset.}
    \label{tab:nyuv2_quant}
\end{table}

\subsection{Indoor data (NYU-Depth-v2)}
\begin{figure*}[htb]
    \centering
    \includegraphics[width=0.95\linewidth]{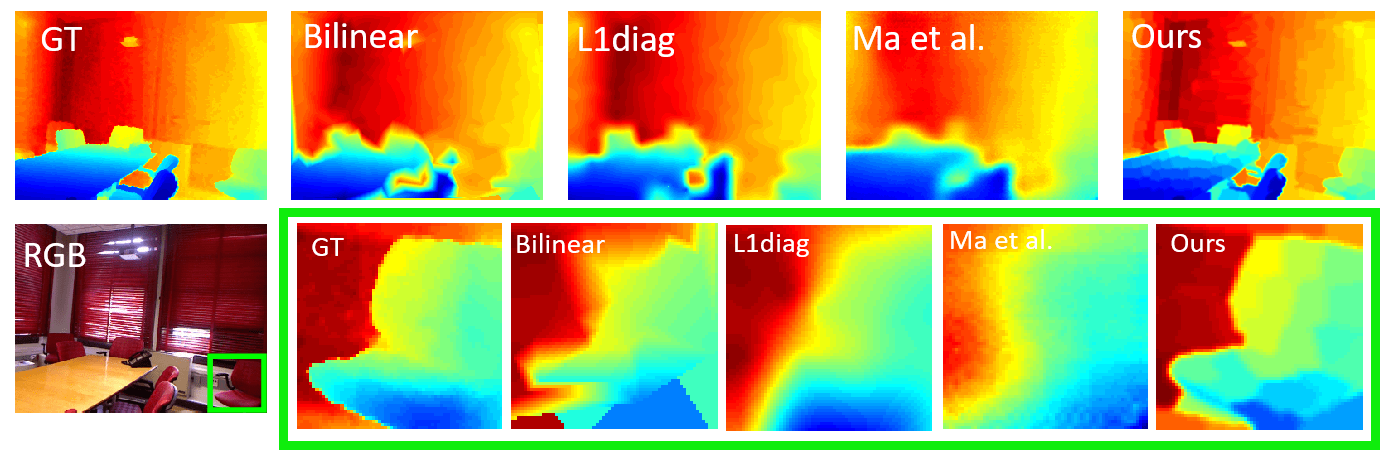}
    \caption{Qualitative results of depth completion on NYU-Depth-v2 dataset at 0.7\% density (500 samples). Tested completion methods: bilinear interpolation, L1diag \cite{ma2017sparse}, Ma et al. \cite{ma2018sparse} and ours.}
    \label{fig:nyuv2_qual}
\end{figure*}

The NYU-Depth-v2 dataset \cite{silberman2012indoor} includes labeled pairs of aligned RGB and dense depth images collected from different indoor scenes. More technical details are given in Section \ref{model_section}.
Quantitative results are listed in Table \ref{tab:nyuv2_quant}. Our method outperforms all other methods. Note that 4 out of the other 6 methods are based on deep learning, while ours is not. 
A qualitative comparison is shown in Fig. \ref{fig:nyuv2_qual}. Although suffering from slight staircase artifacts, our result preserves edges better and stays precise even in small items.

\graphicspath{{./figures/experiments/}}
% NYUv2 - qualitative

\subsection{Prototype: Single-pixel mechanical sampler}
\begin{figure}[htb]
    \centering
    \begin{subfigure}{0.35\textwidth}
        \centering
        \includegraphics[height=1.3in]{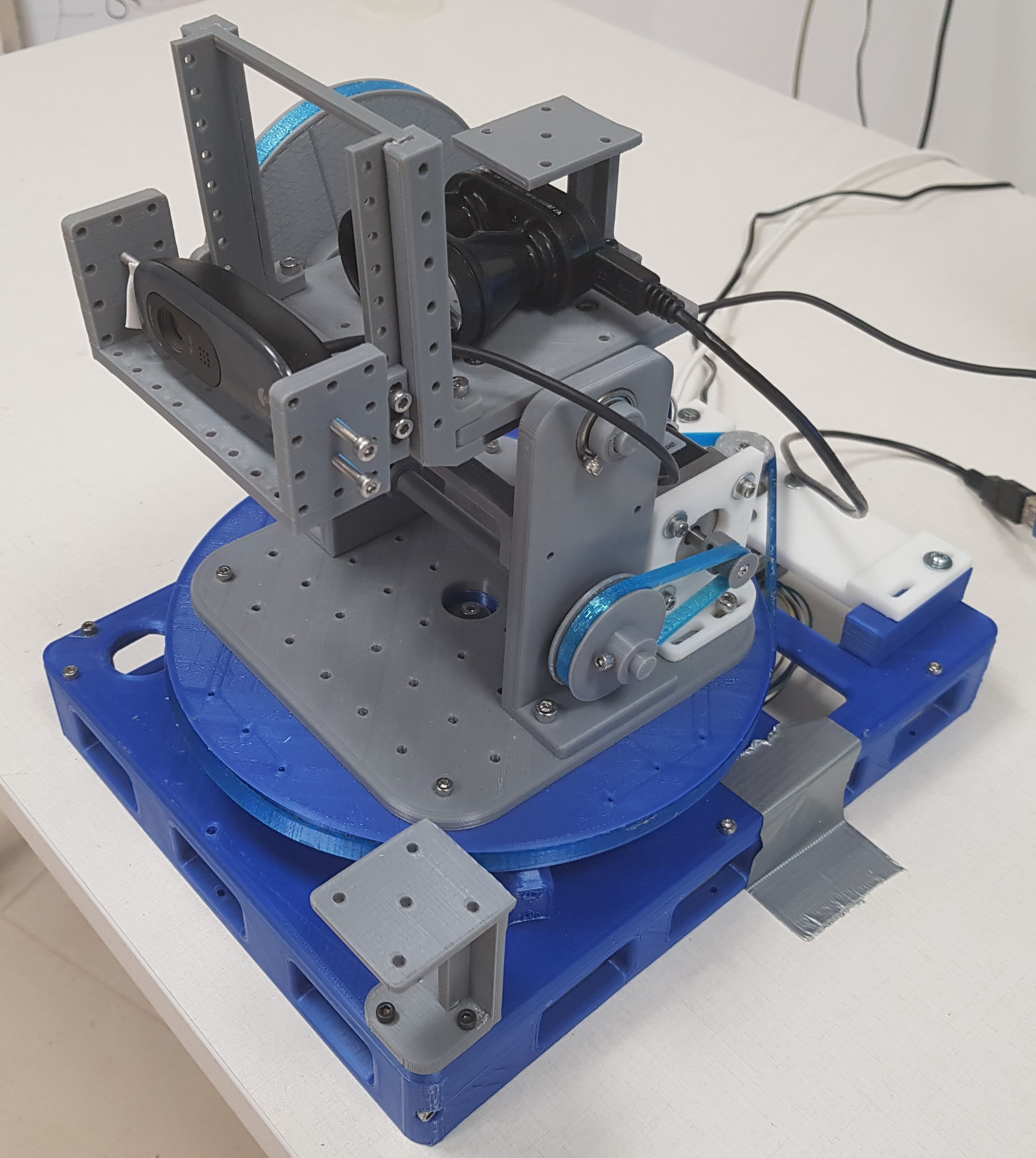}
    \end{subfigure}
    \begin{subfigure}{0.4\textwidth}
        \centering
        \includegraphics[height=1.3in]{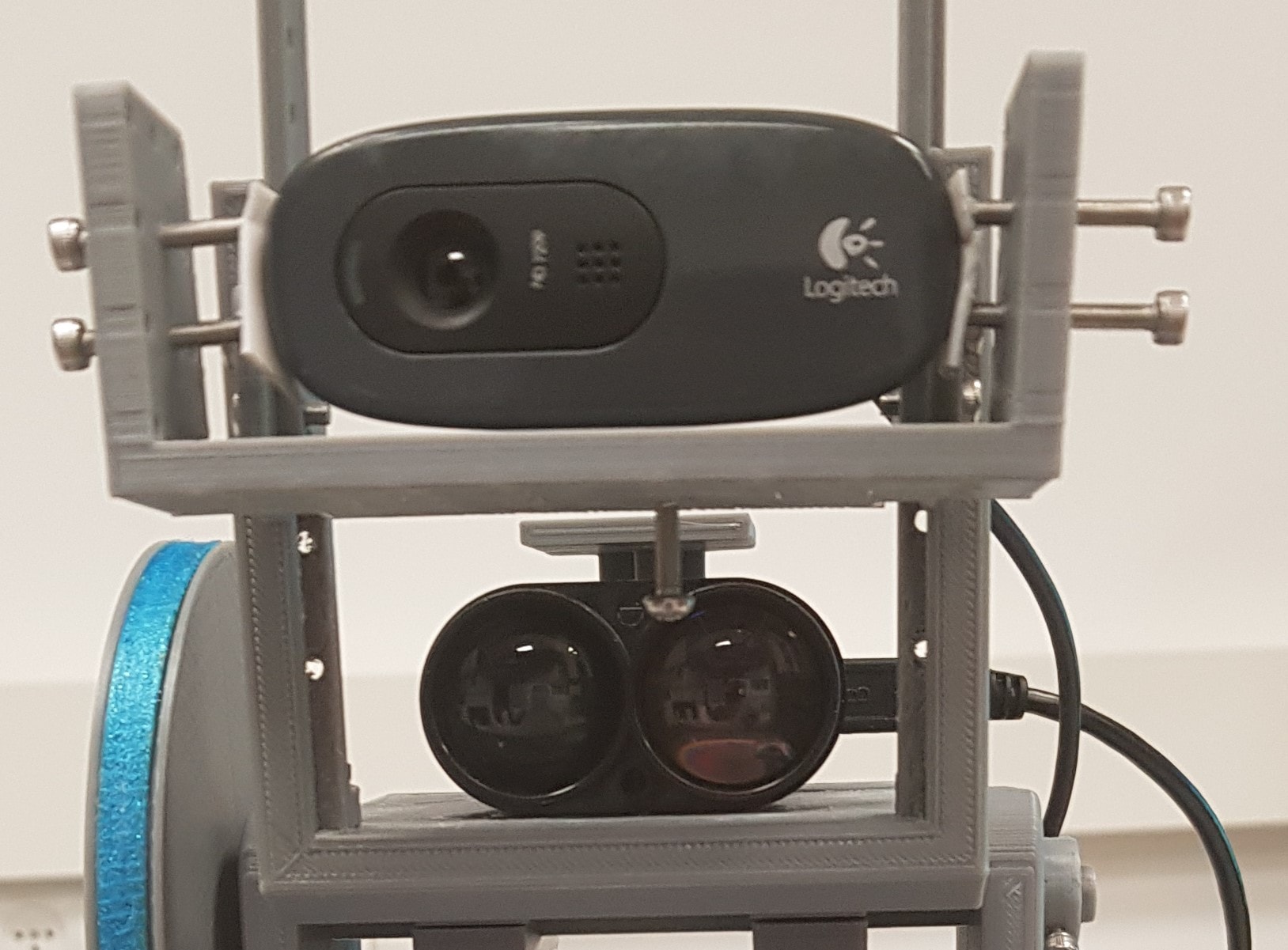}
    \end{subfigure}
    \caption{Our mechanical sampler.}
    \label{fig:device}
\end{figure}

Finally, we performed an experiment over real scene designed in our laboratory. To enable controllable sampling, we built a sampling device (Fig. \ref{fig:device}) assembled by a laser rangefinder, a camera, motors and printed parts. We generated ground-truth images for comparison with Kinect 2 sensor. Note that ground-truth is in real-world coordinate, while our system measures range values.

We created two scenes and sampled them with 3 different patterns to demonstrate the superiority of our method. Results are presented in Fig. \ref{fig:lab_results}. While the first scene (top) is a toy example for testing the sampling resolution, the second scene (bottom) is more realistic. In both cases our method is able to sample all object (even the thinner ones) in the scene and reconstruct them quite accurately.

% lab - device picture
\graphicspath{{./figures/experiments/}}

\newcommand{\rulesep}{\unskip\ \hrule\ }
% lab experiments
\graphicspath{{./figures/experiments/}}
\begin{figure*}[htb]
    \centering
    \begin{subfigure}{0.05\textwidth}
        \centering
        \vspace{70pt}
        \rotatebox{90}{Experiment 1}
    \end{subfigure}
    \begin{subfigure}{0.3\textwidth}
        \centering
        \includegraphics[height=1.0in]{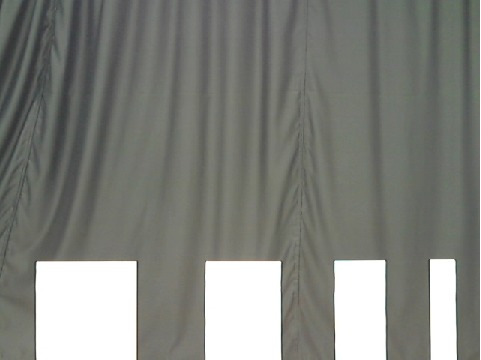}
        \caption{RGB}
    \end{subfigure}
    \begin{subfigure}{0.3\textwidth}
        \centering
        \includegraphics[height=1.0in]{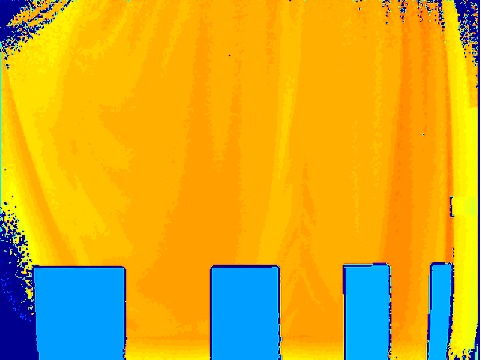} 
        \caption{GT}
    \end{subfigure}
    \begin{subfigure}{0.3\textwidth}
        \centering
        \includegraphics[height=1.0in]{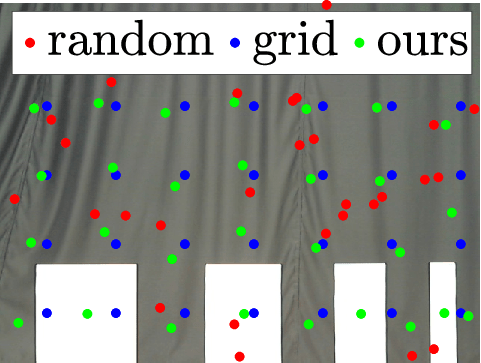} 
        \caption{Depth samples (35 each)}
    \end{subfigure}
    \\
    \begin{subfigure}{0.05\textwidth}
        \centering
        \hspace{0.05pt}
    \end{subfigure}
    \begin{subfigure}{0.3\textwidth}
        \centering
        \includegraphics[height=1.0in]{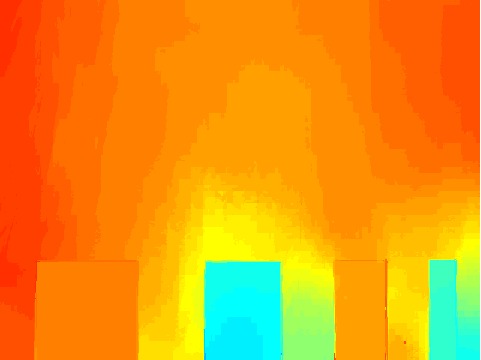} 
        \caption{Random + \cite{barron2016fast}}
    \end{subfigure}
    \begin{subfigure}{0.3\textwidth}
        \centering
        \includegraphics[height=1.0in]{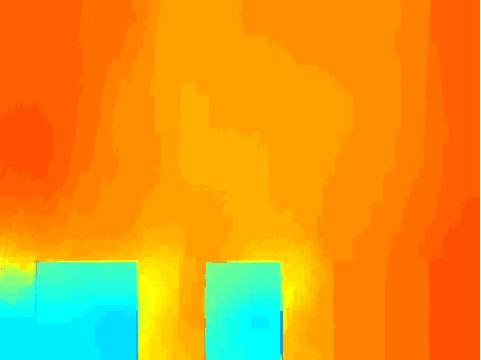} 
        \caption{Grid + \cite{barron2016fast}}
    \end{subfigure}
    \begin{subfigure}{0.3\textwidth}
        \centering
        \includegraphics[height=1.0in]{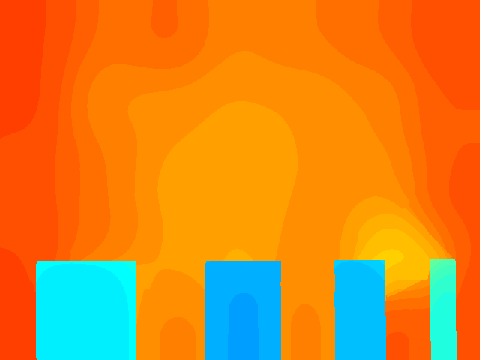} 
        \caption{Ours}
    \end{subfigure}
    \vspace{5pt}
    \rulesep
    \\
    \begin{subfigure}{0.05\textwidth}
        \centering
        \vspace{70pt}
        \rotatebox{90}{Experiment 2}
    \end{subfigure}
    \begin{subfigure}{0.3\textwidth}
        \centering
        \includegraphics[height=1.0in]{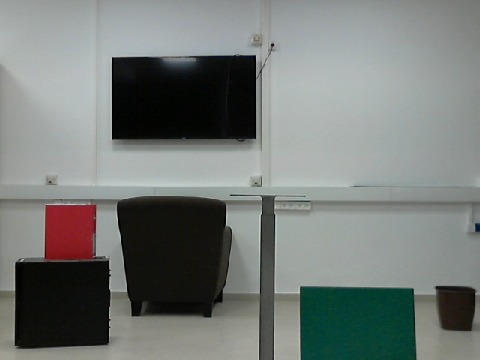} 
        \caption{RGB}
    \end{subfigure}
    \begin{subfigure}{0.3\textwidth}
        \centering
        \includegraphics[height=1.0in]{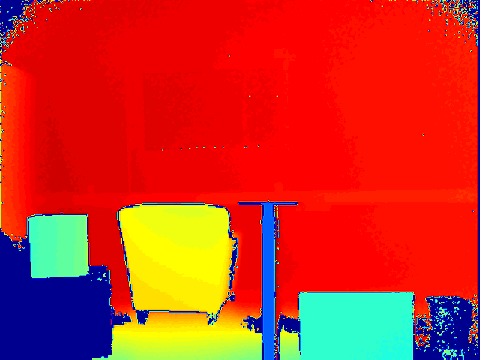} 
        \caption{GT}
    \end{subfigure}
    \begin{subfigure}{0.3\textwidth}
        \centering
        \includegraphics[height=1.0in]{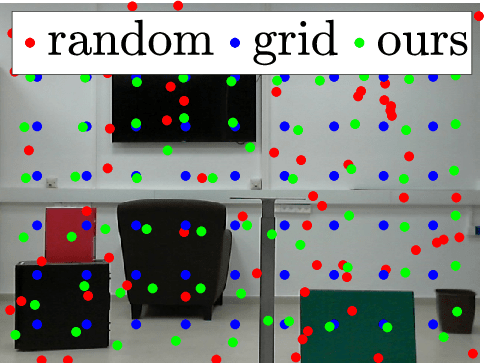} 
        \caption{Depth samples (70 each)}
    \end{subfigure}
    \\
    \begin{subfigure}{0.05\textwidth}
        \centering
        \hspace{0.05pt}
    \end{subfigure}
    \begin{subfigure}{0.3\textwidth}
        \centering
        \includegraphics[height=1.0in]{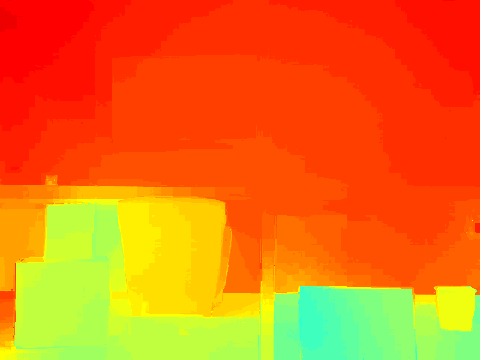}
        \caption{Random + \cite{barron2016fast}}
    \end{subfigure}
    \begin{subfigure}{0.3\textwidth}
        \centering
        \includegraphics[height=1.0in]{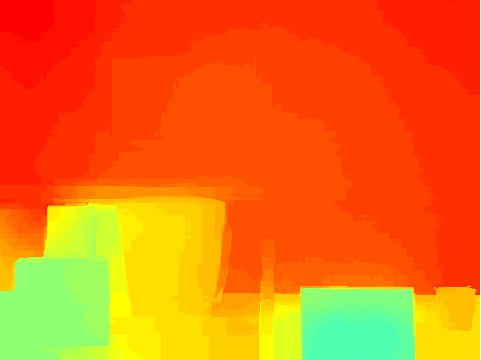}
        \caption{Grid + \cite{barron2016fast}}
    \end{subfigure}
    \begin{subfigure}{0.3\textwidth}
        \centering
        \includegraphics[height=1.0in]{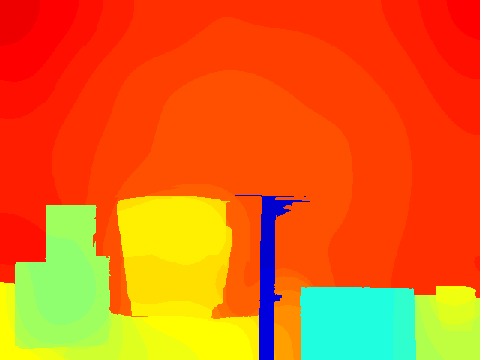} 
        \caption{Ours}
    \end{subfigure}    
    \caption{Experimental results on simple (top) and challenging (bottom) scenes taken in our lab. Random and grid patterns, both reconstructed by bilateral solver \cite{barron2016fast}, fail to sample some of the targets. In contrast, we manage to take a measurement from each object and recover its shape precisely.}
    \label{fig:lab_results}    
\end{figure*}

%-------------------------------------------------------------------------
\clearpage
\newpage
\section{Conclusion}
In this paper, we introduced a novel approach for image-based sparse depth sampling and dense reconstruction. We suggested a parametric piece-wise linear model and have shown its validity for indoor and outdoor datasets. We demonstrated that the correlation between depth and color domains allows to approximate well depth scenes using only an RGB image and a low number of carefully chosen depth samples.
A single-pixel depth sampler was constructed as a proof-of-concept, verifying our predictions.
We believe that this new direction calls for additional extensive research, in order to develop advanced, cheap and accurate depth sensing systems. In future work, we plan to combine classical and modern learning methods to further improve the performance and accuracy.

% \newcommand{\s}{0.15}
% \newcommand{\h}{0.6}
% % Synthia - qualitative
% \graphicspath{{./figures/experiments/}}
% \begin{figure*}[htb]
%     \begin{subfigure}{\s\textwidth}
%         \centering
%         \includegraphics[height=\h\textwidth]{lab_complex/RGB.eps}
%         \caption{RGB}
%     \end{subfigure}
%     \begin{subfigure}{\s\textwidth}
%         \centering
%         \includegraphics[height=\h\textwidth]{lab_complex/depth_gt.eps}
%         \caption{Depth GT}
%     \end{subfigure}
%     \begin{subfigure}{\s\textwidth}
%         \centering
%         \includegraphics[height=\h\textwidth]{lab_complex/samples.eps} 
%         \caption{Depth samples}
%     \end{subfigure}
%     \vspace{10pt}
%     \\
%     \begin{subfigure}{\s\textwidth}
%         \centering
%         \includegraphics[height=\h\textwidth]{lab_complex/rand_recon.eps} 
%         \caption{Random + \cite{barron2016fast}}
%     \end{subfigure}
%     \begin{subfigure}{\s\textwidth}
%         \centering
%         \includegraphics[height=\h\textwidth]{lab_complex/grid_recon.eps} 
%         \caption{Grid + \cite{barron2016fast}}
%     \end{subfigure}
%     \begin{subfigure}{\s\textwidth}
%         \centering
%         \includegraphics[height=\h\textwidth]{lab_complex/our_recon.eps} 
%         \caption{Ours}
%     \end{subfigure}

%     \caption{Additional qualitative results on Synthia dataset with our method (1000 and 5000 samples) and other methods (5000 samples).}
%     \label{fig:lab_results}    
% \end{figure*}

\clearpage
\newpage
{\small
\bibliographystyle{ieee}
\bibliography{ms}
}

\end{document}